\documentclass[journal]{IEEEtran}
\usepackage{subfigure}
\usepackage{caption}
\usepackage{graphicx}
\usepackage{amsthm}
\usepackage{epsfig}
\usepackage{latexsym}
\usepackage{amsfonts}
\usepackage{here}
\usepackage{rawfonts}
\usepackage[latin1]{inputenc}
\usepackage[T1]{fontenc}
\usepackage{calc}
\usepackage{capitalgreekitalic}
\usepackage{url}
\usepackage{enumerate}
\usepackage{color}
\usepackage[tbtags]{amsmath}
\usepackage{amssymb}
\usepackage{upref}
\usepackage{epic,eepic}
\usepackage{times}
\usepackage{dsfont}
\usepackage{comment}
\usepackage{cite}
\usepackage{bbm} 
\usepackage{bm}
\usepackage[caption=false,font=normalsize,labelfont=sf,textfont=sf]{subfig}

\usepackage{dsfont}

\newcounter{step}
\newlength{\totlinewidth}
\newenvironment{algorithm}{%
  \begin{list}{}%
    {\usecounter{step}%
      \settowidth{\labelwidth}{\textbf{Step 2:}}%
      \raggedright}}%
  {\end{list}%
  \rule{\linewidth}{1pt}}
\newcounter{substep}

  {\end{list}}

\IEEEoverridecommandlockouts

\usepackage{algorithm}
\usepackage{algorithmic}

\usepackage{subfigure}

\usepackage{enumitem}

\usepackage{xcolor}
\usepackage{xpatch}
\makeatletter

\usepackage{xcolor}
\usepackage{xpatch}
\makeatletter
\def\changeBibcolor#1{%
\ifin@\color{blue}\else\normalcolor\fi}
\xpatchcmd\@bibitem
{\item}
{\changeBibcolor{#1}\item}
{}{\fai7}
\xpatchcmd\@lbibitem
{\item}
{\changeBibcolor{#2}\item}
{}{\fai7}
\makeatother

\begin{document}
\captionsetup[figure]{labelformat=simple, labelsep=period}
\title{\Huge Optimization of Private Semantic Communication Performance: An Uncooperative Covert Communication Method \vspace*{0.5em}}
\author{{Wenjing Zhang, \emph{Student Member, IEEE}}, Ye Hu, \emph{Member, IEEE}, Tao Luo, \emph{Senior Member, IEEE}, and Zhilong Zhang,  \emph{Member, IEEE}, Mingzhe Chen, \emph{Member, IEEE} \vspace*{-2em}\\ 

\thanks{W. Zhang, T. Luo, and Z. Zhang are with the Beijing Laboratory of Advanced Information Network, Beijing University of Posts and Telecommunications, Beijing, 100876, China (Email: \protect\url{zhangwenjing@bupt.edu.cn}; \protect\url{tluo@bupt.edu.cn}; \protect\url{zhangzhilong@bupt.edu.cn}).}
\thanks{Y. Hu is with the Department of Industrial and Systems Engineering, University of Miami, Coral Gables, FL, 33146 USA (Email: \protect\url{yehu@miami.edu}).}
\thanks{M. Chen is with the Department of Electrical and Computer Engineering and Institute for Data Science and Computing, University of Miami, Coral Gables, FL, 33146 USA (Email: \protect\url{mingzhe.chen@miami.edu}).}
\thanks{This work was supported in part by the National Natural Science Foundation of China under Grant 62171047, U22B2001, and 62271065.}
}

\maketitle
%
\begin{abstract} In this paper, a novel covert semantic communication framework is investigated. Within this framework, a server extracts and transmits the semantic information, i.e., the meaning of image data, to a user over several time slots. An attacker seeks to detect and eavesdrop the semantic transmission to acquire details of the original image. To avoid data meaning being eavesdropped by an attacker, a friendly jammer is deployed to transmit jamming signals to interfere the attacker so as to hide the transmitted semantic information. Meanwhile, the server will strategically select time slots for semantic information transmission. Due to limited energy, the jammer will not communicate with the server and hence the server does not know the transmit power of the jammer. Therefore, the server must jointly optimize the semantic information transmitted at each time slot and the corresponding transmit power to maximize the privacy and the semantic information transmission quality of the user. To solve this problem, we propose a prioritised sampling assisted twin delayed deep deterministic policy gradient algorithm to jointly determine the transmitted semantic information and the transmit power per time slot without the communications between the server and the jammer. Compared to standard reinforcement learning methods, the propose method uses an additional Q network to estimate Q values such that the agent can select the action with a lower Q value from the two Q networks thus avoiding local optimal action selection and estimation bias of Q values. Simulation results show that the proposed algorithm can improve the privacy and the semantic information transmission quality by up to 77.8\% and 14.3\% compared to the traditional reinforcement learning methods. 
\end{abstract}
\begin{IEEEkeywords}
Semantic communication, covert communication, reinforcement learning. 
\end{IEEEkeywords}
\section{Introduction}
 Current communication techniques (e.g., reflected intelligent surface \cite{hua2023ris}, non-terrestrial communications \cite{hu2020leo}, and integrated aerial-ground networks \cite{sun2024multiris}) may not be able to support emerging wireless applications, especially those AI-enabled services, e.g., automatic driving, digital twins, and Metaverse, that require to reliably and efficiently transmit massive volumes of image data that collected by dense visual devices \cite{dusit2022overview,dh2023overview,liu2024aigc}. Semantic communication \cite{Xing2024wcl,jt2024icc, jiahui2024icc,mehdi2024interfer,hjj2024harq,Wangyn2022semantickg} is a novel and promising paradigm to support these resource-intensive services \cite{chaccour2022overview}, by focusing on transmitting only the most relevant meaning of the original data (called \emph{semantic information}) in the receiver's need \cite{gunduz2022overview}. However, in a semantic communication system, the transmitted semantic information that extracted and refined from the original data based on the AI-enabled encoder is much more meaningful \cite{qin2024overview,zp2024overview,wu2023jsccf}, sensitive, and private. A malicious attacker can acquire more valuable information by eavesdropping semantic information in a semantic communication system compared to that in standard communication systems \cite{px2024robustext}. Consequently, the security or privacy issues in semantic communication become more critical \cite{yang2023secure}. Covert communication techniques that can hide the very existence of transmission from a malicious attacker are considered as a powerful solution for transmitting massive image data efficiently and securely. The covert communication technique has been applied widely to secure transmission, e.g., the work in \cite{ma2025covert} optimize the average covert age of information metric by time modulated array.
 However, introducing covert transmission techniques into semantic communication faces several challenges including semantic-related transmit power control, transmission time slot arrangement, evaluation of the semantic information quality, and privacy of the semantic communication system.

\subsection{Related Works}
Recently, several works \cite{zuo2024watermark,lin2023blockchain,Nan2023phl,Ren2024diff,qin2023phl,Liu2023plugmodule,yuhao2023inversion,mj2023wcl} investigated the security issues in semantic communication systems. The authors in \cite{zuo2024watermark} combined semantic coding and digital watermark technique to protect the semantic of the transmitted data from abusing and tampering. The authors in \cite{lin2023blockchain} developed a secure semantic communications framework to prevent data positioning attacks via using blockchains and zero-knowledge proofs. The authors in \cite{Nan2023phl} introduced an adversarial training method to against various semantic-oriented physical adversarial attacks from an imperceptible physical-layer adversarial perturbation generator in semantic transmission. The authors in \cite{Ren2024diff} applied diffusion model to defend semantic-oriented attacks by adversarial purification. However, these works \cite{zuo2024watermark, lin2023blockchain,Nan2023phl,Ren2024diff} do not consider eavesdropping attack, which also poses significant security threats to semantic communications systems. The works in \cite{qin2023phl} introduced physical layer encryption and subcarrier obfuscation solutions to support the security of semantic communication from eavesdropping attack. The work in \cite{Liu2023plugmodule} introduced an unified secure semantic communication framework with three hot-plug-gable semantic protection modules that can secure the transmission by encryption, mitigate leakage risk by perturbation, and calibrate distortion in the receiver by semantic signature generation. The work in \cite{yuhao2023inversion} designed a random permutation and substitution method to prevent the attacker decoding the eavesdropped semantic information. In \cite{mj2023wcl}, the authors introduced a privacy aware loss function that guide the training of joint source and channel coding based neural networks so as to improve the image reconstruction quality while reducing data leakage. However, these encryption and perturbation based methods \cite{qin2023phl,Liu2023plugmodule,yuhao2023inversion,mj2023wcl} focused on the protection of the transmitted information and used the size of protected data as a metric to evaluate protection performance. Hence, they do not consider data meaning eavesdropped by the attacker. In practice, an attacker may not be interested about the entire transmitted data. Hence, considering the data meanings that the attacker is interested can significantly improve the data privacy performance.

Covert communications is a potential solution to secure the wireless transmission with high-security level \cite{chen2023covertsurvey}. Compared to the traditional physical layer security (e.g., hybrid beamforming\cite{zhi2021tcom} and absorptive reconfigurable intelligent surfaces \cite{zhi2024satellite}), covert communication technique can hide the very existence of the wireless transmission without the spectral cost of tedious and redundant encryption management process, by properly controlling transmit power \cite{Bash2013covert} or interfering malicious detection with the help of a friendly jammer \cite{sobers2017jammer}. Hence, some works in \cite{Xu2024crosslayer,hu2024textcovert,wang2023covert, du2024covert} studied the use of covert communication technique in semantic communication system to improve semantic security. In \cite{Xu2024crosslayer}, a covert and reliable semantic communication framework is introduced to against multiple eavesdropping attacks by using a full-duplex receiver that is required to decode the received semantic information and transmit artificial noise simultaneously. Similarly, in \cite{hu2024textcovert}, the full-duplex receiver is also considered to construct a covert semantic communication system. However, these works \cite{Xu2024crosslayer,hu2024textcovert} 
may not be used for energy limited devices since they require to use full-duplex receiver and self-interference concealment schemes, which are energy consuming. In \cite{wang2023covert}, a multi-agent reinforcement learning algorithm is proposed to secure the semantic communication from eavesdropping attack by selectively, and thus temporally, protecting one user with the friendly jammer in the system. Hence, all the rest of users are exposed to the attack's eavesdropping without any protection. In \cite{du2024covert}, a generative diffusion model based covert communication method is developed to hide the semantic transmission. Despite the promising results, these two works \cite{wang2023covert,du2024covert} on covert semantic communication require the connection and cooperation between the transmitter and the jammer, which is cost-expensive and impractical for the most wireless communication scenarios due to the limited energy and computation capacity of the jammer. 

\begin{table}\centering\footnotesize
\setlength{\belowcaptionskip}{0pt}
\setlength{\abovedisplayskip}{-15pt}
\setlength{\tabcolsep}{1.5mm}{
\newcommand{\tabincell}[2]{\begin{tabular}{@{}#1@{}}#2.0\end{tabular}}
\renewcommand\arraystretch{1.5}
\caption[table]{{List of notations}}
\label{table:notations}
\centering
\begin{tabular}{|c|c|}
\hline
\!\textbf{Notation}\! \!\!& \!\!\!\!\textbf{Description} \\
\hline
$n$ & Index of transmission time slot \\
\hline
$N$ & Number of transmission time slots \\
\hline
$B$ & Number of triple transmitted at one time slot \\
\hline
$W$ & Bandwidth for transmission     \\
\hline
$p_J^n$ & Transmit power of the jammer \\
\hline
$p_S^n$ & Transmit power of the server \\
\hline
$\beta$ & Path loss exponent \\
\hline
$\sigma_U$ & Standard deviation of Gaussian noise\\
\hline
$\bm{q}^n$ & triple transmission vector\\
\hline
$\zeta_U^n$ & Power of the received signal \\
\hline
$c_U^n$ & Downlink data rate\\
\hline
$t_U^n$ &Transmission latency\\
\hline
$T$ & Transmission latency threshold\\
\hline
$\bm{\Psi}$ & Semantic information\\
\hline
$\bm{\psi}^k$ & Semantic triple $k$ in $\bm{\Psi}$ \\
\hline
$Z\left(\bm{\psi}^k\right)$ & Number of bits to transmit $\bm{\psi}^k$ \\
 \hline
 $\bm{g}$ & Channel monitoring vector \\
\hline
$\epsilon_n$ & Power detection threshold \\
\hline
$C\left(\bm{\psi}^k\right)$ & Vectorized semantic triple $\bm{\psi}^k$  \\
\hline
$E_U$ & Semantic similarity metric\\
\hline
$G$ & Maximum detection number \\
\hline
$\gamma$ & Semantic privacy threshold \\
\hline
$\eta$ & Penalty of insecure transmission \\
\hline
\end{tabular}}
\end{table}

\subsection{Contributions}
The main contribution of this work is a novel covert semantic communication framework that enables the server to consistently and privately transmit semantic information to the user against eavesdropping attacks without information sharing with a friendly jammer. The key contributions include:
\begin{itemize}

\item We propose a novel covert semantic communication framework within which a server transmits image data to a user utilizing semantic communication techniques, while a friendly jammer is deployed to protect this user from multiple eavesdropping attacks during semantic transmission process by transmitting interference jamming signals. The server and friendly jammer
cannot communicate or coordinate with each other such that the server transmits semantic information without the help of channel state information or the transmit power of the jammer. Therefore, the jammer in the proposed framework is also not required to estimate wireless channel and chooses jamming signal power independently. To improve the quality of the received semantic information of the user and system privacy, the server must jointly optimize the semantic information transmitted at each time slot and the corresponding transmit power without aid of connection and coordination from the jammer.
\item To evaluate the quality of the received semantic information of the user and privacy performance of the semantic transmission, a novel metric called graph-to-nearest-triple (GNT) is introduced. This metric can directly capture the correlation of the meanings of the extracted and received semantic information, considering the image contents rather than bit errors, and does not require access to the embedding of the original image.
\item We formulate this semantic information selection and transmit power control problem as an optimization problem whose goal is to maximize the semantic similarity of the user while improve system privacy, i.e., guaranteeing the semantic similarity of the attacker being low. Since the optimization problem is non-convex and the objective function and semantic communication privacy constraint depends on graph embedding neural network model, the relationship between the objective function and optimization variables cannot be represented exactly. In consequence, such a maximization problem cannot be solved by traditional optimization methods. To solve the problem, we introduce a prioritized sampling assisted deep RL method that can be trained offline with less interactions between the server and the user such that reducing training data collection overhead.

\item To prevent the RL algorithm converging to sub-optimal policies resulting from Q values overestimation, we introduce a clipped double Q learning technique that can learn Q value function in a conservative and steady way. Meanwhile, to improve the convergence speed of this delayed RL, we also introduce a prioritized sampling mechanism that enables the agent to learn from more valuable and unexpected experience by including the sampling priority in historical experience replaying in training stage.
\end{itemize}

The rest of this paper is organized as follows. The proposed covert image semantic communication system model and the problem formulation are described in Section \uppercase\expandafter{\romannumeral2}. Section \uppercase\expandafter{\romannumeral3} introduces the proposed prioritized sampling assisted twin delayed deep deterministic policy gradient algorithm for private semantic transmission optimization. In Section \uppercase\expandafter{\romannumeral4}, numerical results are presented and discussed. Finally, conclusion are drawn in Section \uppercase\expandafter{\romannumeral5}.

\section{System Model and Problem Formulation}\label{se:system}
\begin{figure}
\centering
\setlength{\belowcaptionskip}{-0.05cm}
\includegraphics[width=7cm]{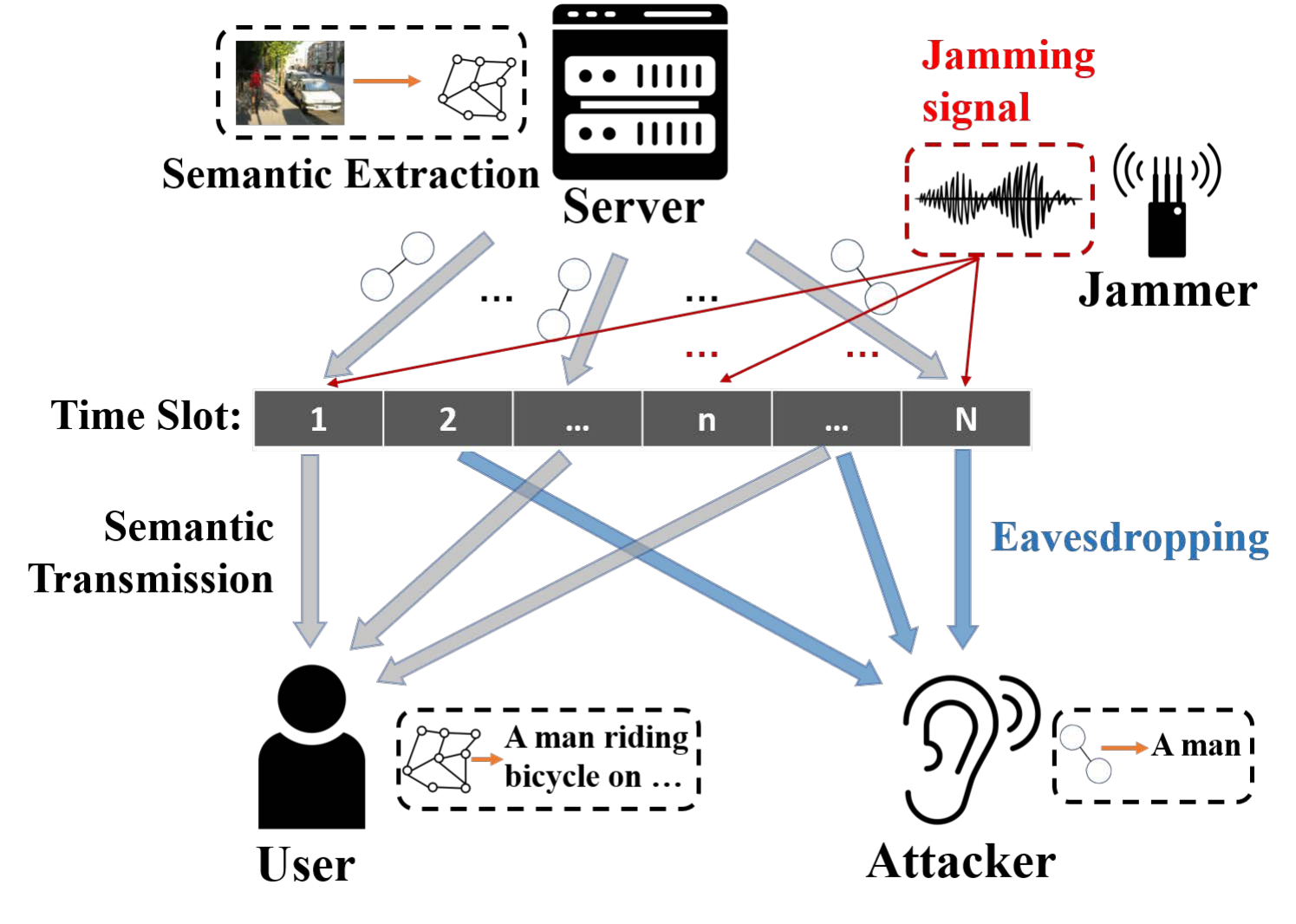}
\caption{Secure semantic communication wireless network.}
\label{fig1}
\vspace{-0.2cm}
\end{figure}

Consider a cellular network in which a server transmits the meaning of images to a user using semantic communication techniques while avoiding detection by an eavesdropping attacker. In the considered model, as shown in Fig.~\ref{fig1}, the server will determine whether to transmit semantic information at each time slot. We assume that the server needs to use $N$ time slots to complete semantic information transmission. Meanwhile, the user can receive the semantic information transmitted at any time slots. The attacker can only select $G$ time slots from the $N$ time slots to eavesdrop data due to its limited energy. To achieve private semantic transmission based on covert communication technique, a friendly jammer is used to protect the semantic information transmitted by the server via transmitting jamming signals that can interfere the attacker who cannot acquire the meaning of the transmitted image precisely from the eavesdropped semantic information. Furthermore, the jammer cannot communicate with the server or the user such that the jammer does not know the transmit power of the server while the server does not know the power of the attacker transmitting a jamming signal at each time slot. Next, we first introduce the procedure of the semantic information extraction. Then, we present the private semantic information transmission. Finally, we define a metric to evaluate the quality of the received semantic information and introduce the problem formulation. Table \ref{table:notations} summarizes all parameters used in our
work.

\subsection{Semantic Information Extraction}
\begin{figure}
\centering
\setlength{\belowcaptionskip}{-0.05cm}
\includegraphics[width=6.5cm]{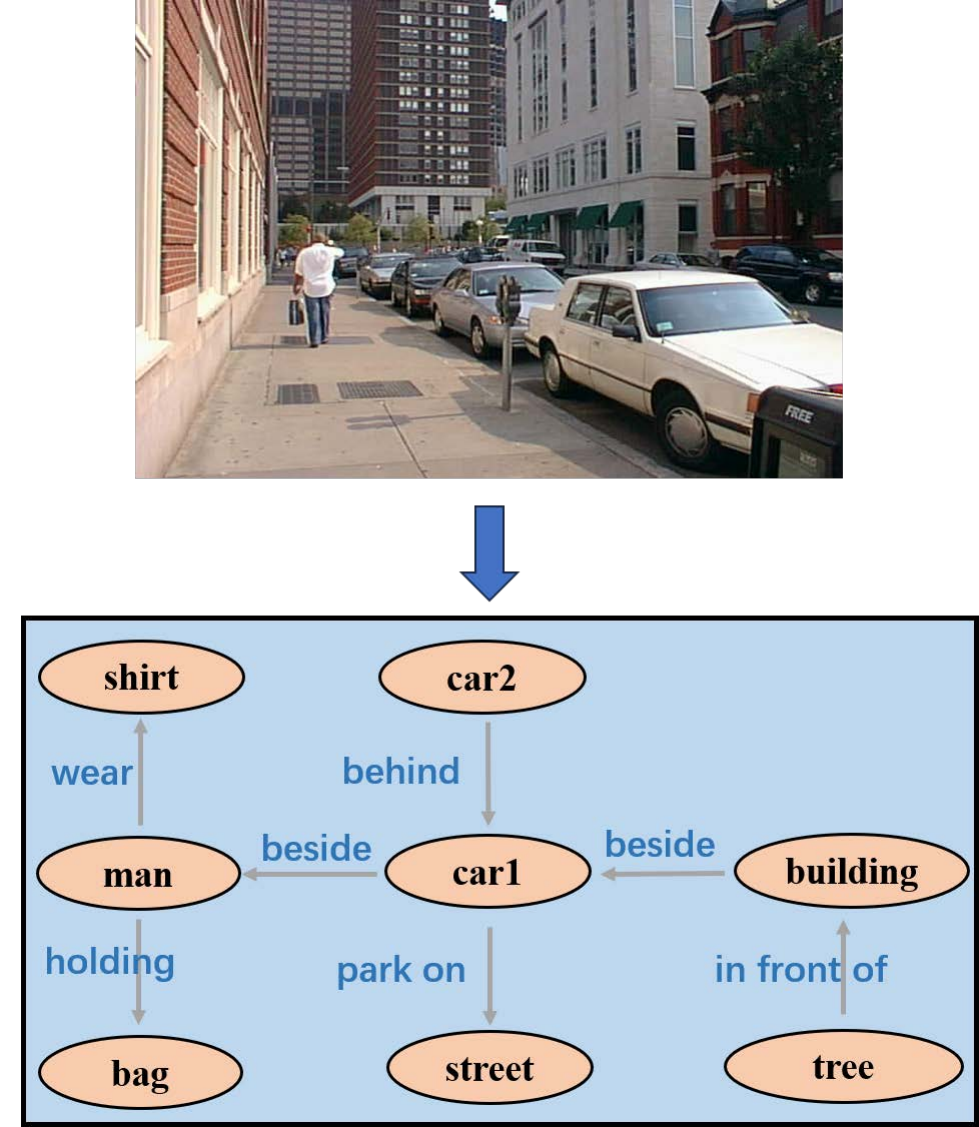}
\caption{Semantic information extraction.}
\label{fig2}
\vspace{-0.2cm}
\end{figure}

We consider image semantic information transmission scenarios, in which a human receiver who wants to acquire the meaning of the transmitted image without the need of image reconstruction. Therefore, we model the semantic information of the image in an explainable form, i.e., a scene graph that consists of several nodes and edges, where a node represents an object and an edge represents the relationship between two objects. Hence, the semantic information consists of both the objects and their relationships in the image. We define a basic component of semantic information, called semantic triple. Each triple consists of two objects and the relationship between them. For example, as shown in Fig.~\ref{fig2}, a semantic triple is ([``\emph{man}''], [``\emph{holding}''], [``\emph{bag}'']), where [``\emph{holding}''] is the relationship between objects [``\emph{man}''] and [``\emph{bag}'']. Such graph-based semantic information enables the human receiver to obtain a comprehensive understanding of the image, e.g., the objects and their relationships in the image. The considered graph based semantic information can be applied for diverse emerging applications such as robotics \cite{robotics}, augmented reality \cite{ar}. In addition, it can also be integrated with the knowledge graph \cite{kg}.


To obtain the semantic information from the original image, a scene graph extraction model in \cite{TDE} is used. Specifically, the server firstly uses the model to detect, locate, and categorize all the objects in the original image. Then, the relationships between all the objects are deduced and represented in form of triples. The extracted semantic information is
\begin{equation}
    \bm{\Psi}=\left\{\bm{\psi}^1,\bm{\psi}^2,\ldots,\bm{\psi}^k,\ldots,\bm{\psi}^K\right\},
\vspace{-0.1cm}
\end{equation}
where $K$ is the number of semantic triples in image and $\bm{\psi}^k=\left(e_k^i,l_k,e_k^j\right)$ is a semantic triple with $l_k$ being the relationship between objects $e_k^i$ and $e_k^j$.

\subsection{Covert Communication Based Private Semantic Transmission}
 We assume that the server transmits $K$ semantic triples extracted from the original image via $N$ transmission time slots. {Here, no specific constraint is imposed on the relationship between $K$ and $N$, i.e., the both $K\geqslant N$ and $K\leqslant N$ are considered possible.}
 Specifically, a triple transmission matrix over $N$ time slots can be given by
\begin{equation}
\bm{Q}=\left[\bm{q}_1,\bm{q}_2,\ldots,\bm{q}_n,\ldots,\bm{q}_N\right],  \end{equation}
 where $\bm{q}^n=\left[q_1^n,q_2^n,\ldots,q_k^n,\ldots,q_K^n\right]$ is triple transmission vector at time slot $n$ with $q_k^n=1$ representing that the server will transmit semantic triple $\psi^k$ at slot $n$, and $q_k^n=0$, otherwise. We assume that the semantic triples are only transmitted over AWGN channels once, i.e., $\sum_{n=1}^Nq_k^n=1$. Hence, when $K\geqslant N$, the server will choose a subset of semantic triples to transmit and decides the transmit power at each time slot. When $K < N$, the server has more time slots for semantic triple transmission, thus improving system privacy level.
 Then, the power of the received signal at the user $\zeta_U^n$ and attacker $\zeta_A^n$ at time slot $n$ can be given as 
\begin{equation}
\zeta_U^n=p_S^nd_{S,U}^{-\beta}+p_J^nd_{J,U}^{-\beta}+\sigma_U^2   
\end{equation}
and 
\begin{equation}
\zeta_A^n=p_S^n{d_{S,A}^{-\beta}}+p_J^nd_{J,A}^{-\beta}+\sigma_A^2,     
\end{equation}
where $p_S^n$ and $p_J^n$ are the transmit power of the server and the jammer at time slot $n$, respectively, $d_{S,U}$ and $d_{S,A}$ are the distance from the server to the user and the attacker, $d_{J,U}$ and $d_{J,A}$ are the distance from the jammer to the user and the attacker, $\beta$ is the path loss exponent, $\sigma_U^n$ and $\sigma_A^n$ are the variance of the Gaussian noise at the user and the attacker, respectively. We consider the transmit power $p_J^n$ of the jammer at each time slot $n$ follows an uniformed distribution with a range $\left[0, p_J^{\text{max}}\right]$.

The data rate of the server transmitting a semantic triple to the user at time slot $n$ can be given as 
\begin{equation}
    c_U^n=W\log\left(1+\frac{p_S^nd_{S,U}^{-\beta}}{p_J^nd_{J,U}^{-\beta}+\sigma_U^2}\right),
\end{equation}
where $W$ is the bandwidth. Then, the transmission latency of semantic triple $\bm{\psi}^k$ at time slot $n$ is $t_U^n =\frac{Z\left(\bm{\psi}^k\right)}{c_U^n}$, where function $Z\left(\bm{\psi}^k\right)$ is the number of bits that the server requires to transmit $\bm{\psi}^k$ over wireless links. Similarly, we can obtain the transmission data rate and latency of the attacker as $c_A^n$ and $t_A^n$. 

In the considered system, the semantic transmission fails and the received incomplete semantic triple will be discarded when transmission latency is larger than the predefined latency threshold $T$. The user will not ask the server to resend the discarded semantic information. This assumption is used to guarantee the minimum triple transmission speed. Therefore, the received semantic information at the user is 
\begin{equation}
\bm{\Psi}_U\left(\bm{p}_S,\bm{p}_J,\bm{Q}\right)=\left\{\bm{\psi}_U^1,\bm{\psi}_U^2,\ldots,\bm{\psi}_U^k,\ldots,\bm{\psi}_U^K\right\},  
\end{equation}
where $\bm{\psi}_U^k=\bm{\psi}^k$ is the received semantic triple if transmission latency satisfying $t_U^n\leqslant T$. Otherwise, $\bm{\psi}_U^k$ is null. $\bm{p}_S=\left[p_S^1,\ldots,p_S^n,\ldots,p_S^N\right]$
and
$
\bm{p}_J=\left[p_J^1,\ldots,p_J^n,\ldots,p_J^N\right]  $
are transmit power vector of the server and the jammer. 

\subsection{Detection and Eavesdropping of the Attacker}
In our work, we assume that the attacker will only monitor the channel in a limited number of time slots since 1) the attacker has limited energy and 2) the attacker may monitor other channels. Hence, the attacker will know the received power of the transmitter within limited time slots. We define a channel monitoring vector $\bm{g}=\left[g_1,g_2,\ldots,g_n,\ldots,g_N\right]$ which indicates the time slots that the attacker selected to monitor the channel and obtain the received power of the transmitter, $g_n=1$ indicates that the attacker chooses to monitor the channel at time slot $n$. In contrast, $g_n=0$ represents that the attacker will not monitor the channel during that slot. When the attacker monitors the channel, a radiometer method \cite{theoreticsecurity,powerthreshold} will be applied to assess the presence of a transmission based on received power of the transmitter. In particular, the attacker compares the total received power with a power threshold $\epsilon_n$. The detection result at time slot $n$ is 
\begin{equation}\label{powerdetector}
D_n=\left\{
\begin{aligned}
 1,\quad &\zeta_A^n\geqslant \epsilon_n,\\
0,\quad &\zeta_A^n< \epsilon_n.
\end{aligned}
\right.
\end{equation}
If the power detector identifies the presence of data transmission over the channel, i.e.,  $D_n=1$, the attacker will detect and decode the received signal to acquire transmitted semantic triple. From (\ref{powerdetector}), we see that the attacker cannot decode a semantic triple if the received signal power is larger than the power threshold while the server does not transmit any triples.
  
 The semantic information eavesdropped by the attack is 
\begin{equation}
\bm{\Psi}_A\left(\bm{p}_S,\bm{p}_J,\bm{Q},\bm{g}\right)=\left\{\bm{\psi}_A^1,\bm{\psi}_A^2,\ldots,\bm{\psi}_A^k,\ldots,\bm{\psi}_A^K\right\},  
\end{equation}
where $\bm{\psi}_A^k=\bm{\psi^k}$ is the eavesdropped semantic triple if $q_k^ng_nD_n\mathbbm{1}_{\left\{t_A^n\leqslant T\right\}}=1$ holds. Otherwise, $\bm{\psi}_A^k$ is null. We can see that the attacker successfully eavesdrops the transmission of the triple $\bm{\psi}_k$ at time slot $n$ ($q_k^n=1$) only when 1) the attacker selects time slot $n$ to detect (i.e., $g_n=1$), 2) the attacker successfully detect the semantic triple (i.e., $D_n=1$), 3) the transmission latency $t_A^n$ from the server to the attacker is smaller than the latency threshold $T$ (i.e., $\mathbbm{1}_{\left\{t_A^n\leqslant T\right\}}=1$). Therefore, to prevent the attacker from eavesdropping the transmitted semantic information, the server have to jointly optimize transmit power and select triples to transmit at each time slot.
\begin{figure}
\centering
\setlength{\belowcaptionskip}{-0.05cm}
\includegraphics[width=9cm]{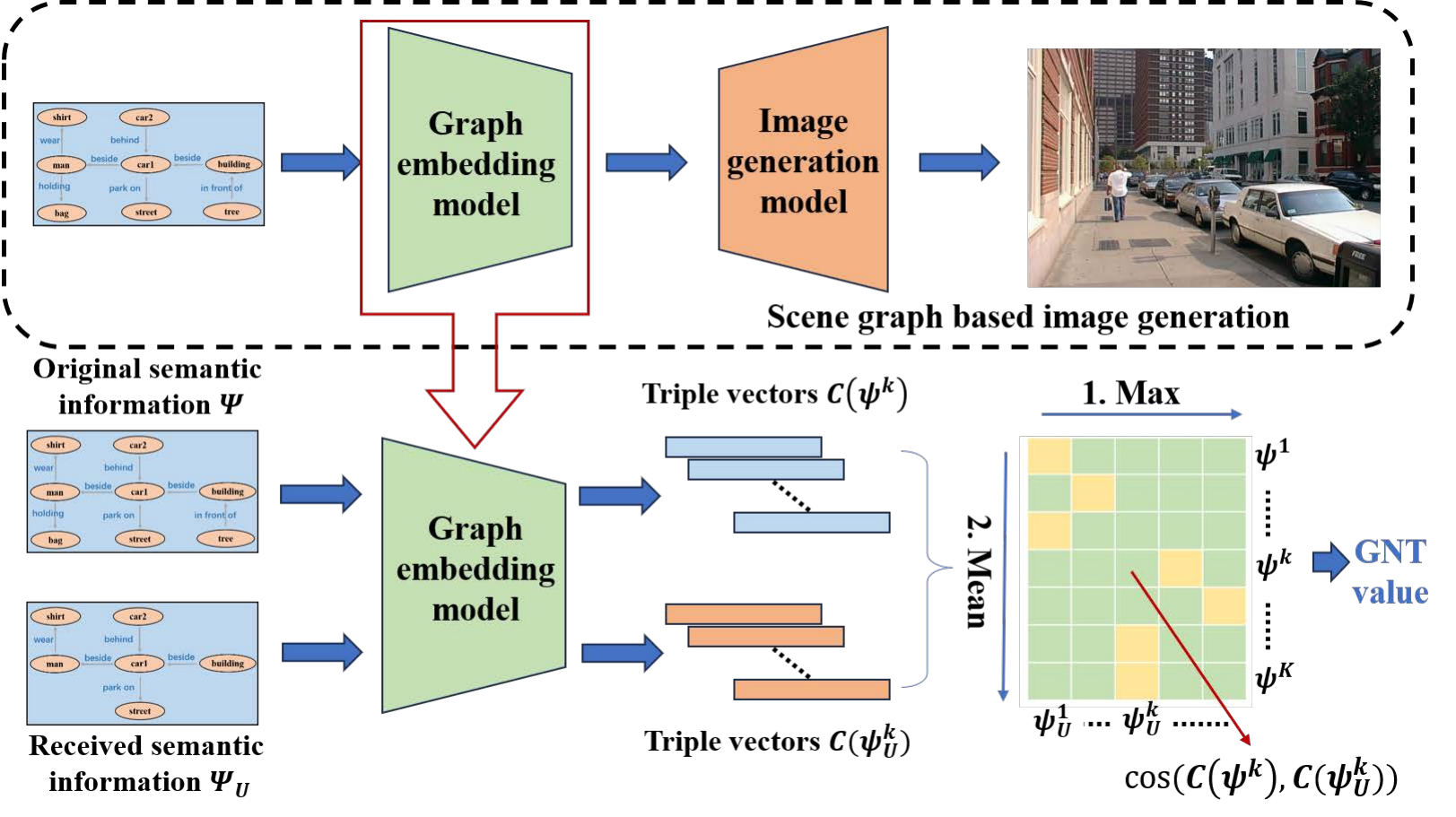}
\caption{Image-to-nearest-triple metric for semantic information.}
\label{fig3}
\vspace{-0.2cm}
\end{figure}
\subsection{Metric for Semantic Communication}
 To evaluate the quality of the received semantic information and the eavesdropped semantic information, we cannot use mean square errors between the original image and the received data since there is no image reconstruction in the considered system. To this end, we introduce a metric called graph-to-nearest-triple (GNT). Different from the metric in \cite{ISS} that measures the similarity between the extracted semantic information and the original image, the proposed GNT can capture the correlation of the meanings of the extracted and received semantic triples by directly comparing the image contents, thus, exclude the need for accessing the embedding of the original image. 
 
 The proposed metric accounts both the number and importance of received semantic triples. Specifically, the proposed GNT metric of the user is given by 
\begin{equation}\label{GNT}
E_U\left(\bm{p}_S,\bm{p}_J,\bm{Q}\right)=\frac{1}{K}\sum\limits_{k=1}^{K}\mathop{\max}\limits_{\bm{\psi}_U^k}\left\{\cos\left(C\left(\bm{\psi}^k\right),C\left(\bm{\psi}_U^k\right)\right)\right\},
\end{equation}
where $\bm{p}_S$ and $\bm{p}_J$ are the transmit power vector of the server and the jammer, $\bm{Q}$ is the triple transmission vector, $\bm{\psi}^k$ and $\bm{\psi}^k_U$ are the original triple $k$ and the received triple $k$ at the user, and $C\left(\cdot\right)$ is the pre-trained scene graph embedding model. Compared to the common embedding models such as BERT, the used graph embedding model accounts the semantic relationships between objects by messaging passing. In addition, the used embedding model can also encode the input scene graph while considering image semantics. The cosine similarity between the original triple $\bm{\psi}^k$ and the received triple ${\psi}_U^k$ is $\cos\left(C\left(\bm{\psi}^k\right),C\left(\bm{\psi}_U^k\right)\right)=\frac{C\left(\bm{\psi}^k\right)C\left(\bm{\psi}_U^k\right)^T}{\Vert C\left(\bm{\psi}^k\right)\Vert\Vert C\left(\bm{\psi}_U^k\right)\Vert}$. Here, we note that $C\left(\bm{\psi}_U^k\right)=\bm{0}$ if $\bm{\psi}_U^k$ is null, which implies that this triple is not received by user $k$ due to the high transmission latency. 

From (\ref{GNT}), we can see that as the number of received triples increases, the GNT value will increase. Meanwhile, the designed metric can measure the importance of semantic triples. For example, let $\bm{\psi}^*$ be the key triple and hence, it is different from other triples. If $\bm{\psi}^*$ is discarded, the GNT value of the user will be significantly decreased from $\cos\left(C\left(\bm{\psi}^*\right),C\left(\bm{\psi}^*\right)\right)=1$ to $\mathop{\max}\limits_{k\neq*}\left\{\cos\left(C\left(\bm{\psi}^k\right),C\left(\bm{\psi}^*\right)\right)\right\}$. However, if a common triple is lost, we can still find its similar triple to obtain high cosine similarity thus achieving high GNT value. Therefore, the designed GNT can measure the importance of semantic triples.

Similarly, the GNT of the eavesdropped semantic information $\bm{\Psi}_A\left(\bm{p}_S,\bm{p}_J,\bm{Q},\bm{g}\right)$ is 
\begin{equation}
E_A\left(\bm{p}_S,\bm{p}_J,\bm{Q},\bm{g}\right)=\frac{1}{K}\sum\limits_{k=1}^{K}\mathop{\max}\limits_{\bm{\psi}_A^k}\left\{\cos\left(C\left(\bm{\psi}^k\right),C\left(\bm{\psi}_A^k\right)\right)\right\}.
\end{equation}

\subsection{Problem Formulation}
Given the defined system model, our objective is to maximize the semantic similarity of the user while ensuring the semantic similarity of the attacker being low. This maximization problem optimizes the triple transmission matrix $\bm{Q}$ and the transmit power vector $\bm{p}_S$ of the server, which is formulated as 
\begin{subequations}\label{eq:optimal_problem}
\begin{align}\tag{11}
&{\mathop{\max}\limits_{\bm{p}_{S},\bm{Q}}E_U\left(\bm{p}_S,\bm{p}_{J},\bm{Q}\right)-E_A\left(\bm{p}_S,\bm{p}_J,\bm{Q},\bm{g}\right)}\\
&{\;\;{\rm{s}}.{\rm{t}}.\;\;\;\;p_S^n\leqslant p_{S}^{\text{max}}},\\
&{\;\;\;\;\;\;\;\;\;\;\;p_J^n\sim U(0,p_{J}^{\text{max}}}),\\
&{\;\;\;\;\;\;\;\;\;\;\;\sum_{k=1}^K q_k^n\leqslant B, \sum_{n=1}^N q_k^n=1,}\\
&{\;\;\;\;\;\;\;\;\;\;\;\sum_{n=1}^N g_n\leqslant G,}\\
&{\;\;\;\;\;\;\;\;\;\;\;E_A\left(\bm{p}_S,\bm{p}_J,\bm{Q},\bm{g}\right)\leqslant \gamma},
\end{align}
\end{subequations}
 where $p_S^{\text{max}}$ and $p_J^{\text{max}}$ are the maximal transmit power of the server and the jammer, respectively. Constraint (\ref{eq:optimal_problem}a) limits the transmit power of the server at each time slot. (\ref{eq:optimal_problem}b) indicates that the jamming signal power follows an uniformed distribution. Constraint (\ref{eq:optimal_problem}c) limits the number of triples that can be transmitted per time slot and each triple can only be transmitted once. Since only partial time slots are used for semantic transmission, it is possible that some slots are not used, i.e., $\sum_{k=1}^Kq_k^n=0$. (\ref{eq:optimal_problem}d) implies that the attacker can detect semantic information at most $G$ time slots. Constraint (\ref{eq:optimal_problem}e) is the semantic communication privacy requirement. From (\ref{eq:optimal_problem}), we can see that the proposed optimization problem is non-convex and the objective function and constraint (\ref{eq:optimal_problem}e) depends on scene graph embedding neural network model. Hence, the relationship between the objective function and optimization variables cannot be represented exactly. In consequence, the problem (\ref{eq:optimal_problem}) cannot be solved by traditional optimization algorithms.

\section{Prioritized Sampling Assisted Twin Delayed Deep Deterministic Policy Gradient Algorithm}\label{se:system}
 We introduce a prioritized sampling assisted twin delayed deep deterministic policy gradient algorithm (PS-TD3) to solve the problem in (\ref{eq:optimal_problem}). Compared to current RL methods \cite{ddpg}, the proposed algorithm can accurately estimate Q value function by clipped double Q learning so as to stably improve semantic quality of the received semantic information of the user and system privacy. Moreover, the introduced prioritized sampling technique can enable the agent to learn more from surprising and unexpected historical experience such that the training speed of the proposed method is improved. Next, we will first introduce the components of the proposed algorithm, and then explain the training process of the algorithm.
 
\subsection{Components of PS-TD3 Algorithm}
In this section, we introduce the fundamental components of the proposed PS-TD3 algorithm as follows:
\begin{itemize}
\item{\emph{Agent}}: The agent is the server that consecutively decides the triple to transmit, and power that is used for this transmission, at each time slot. 

\item{\emph{States}}: The state captures the mutual importance distribution of all semantic triples, such that the state at time slot $n$ is defined as a vector $\bm{s}_n=\left[\bm{\xi}^n_1,\ldots, \bm{\xi}^n_k,\ldots,\bm{\xi}^n_K\right]$, with $\bm{\xi}^n_k=\left[\cos\left(C\left(\bm{\psi}^k\right),C\left(\bm{\psi}^1\right)\right),\ldots,\cos\left(C\left(\bm{\psi}^k\right),C\left(\bm{\psi}^K\right)\right)\right]$ being the importance distribution of semantic triple $\bm{\psi}^k$ at time slot $n$. The sequence of server states recorded until the completion of semantic triple transmission is captured by a vector $\left[\bm{s}_1,\ldots,\bm{s}_n,\ldots,\bm{s}_N\right]$. Based on such mutual importance distribution, the server can find higher important triples that with higher cosine similarity for the most of triples. Hence, the server will put higher priority of this important triples in transmission so as to improve communication system privacy. Meanwhile, since the server does not know the transmit power of jamming signals, the server state does not include the transmit power of jamming signals.

\item{\emph{Actions}}: The action of server consists of choosing partial semantic triples to transmit and determining the corresponding transmit power at each time slot. In particular, at time slot $n$, the server action is $\bm{a}_n=\left(\bm{q}^n,p_S^n\right)$, with $\bm{q}^n$ being the transmission vector and $p_S^n$ being the corresponding transmit power. The sequence of actions select over the whole semantic triple transmission process is $\left[\bm{a}_1,\ldots,\bm{a}_n,\ldots,\bm{a}_N\right]$

\item{\emph{Reward}}: The reward of the server capture the benefits of a selected action in terms of semantic
transmission quality and system privacy. To achieve higher reward, the agent is required to maximize semantic similarity of the user and meanwhile minimize semantic similarity of the attacker, such that the reward at each step $n$ is
\begin{equation}\label{tmp reward}
r_n=\left(E_U^n-E_U^{n-1}-E_A^n+E_A^{n-1}\right)\mathbbm{1}_{\left\{E_A^n\leqslant\gamma\right\}},
\end{equation}
where $E_U^n$ is the temporal GNT of the received semantic information at the user, $E_A^n$ is the temporal GNT of the eavesdropped semantic information, at the $n$-th time slot. $\mathbbm{1}_{\left\{E_{A}^n\leqslant\gamma\right\}}$ is the indicator of the system privacy level, with $\mathbbm{1}_{\left\{E_{A}^n\leqslant\gamma\right\}}=1$ implying that the semantic transmission is private at time slot $n$, and $\mathbbm{1}_{\left\{E_{A}^n\leqslant\gamma\right\}}=0$, otherwise. From (\ref{tmp reward}), we see that the temporal reward will be zero with under-threshold privacy performance. 

The sequence of state-action-reward transition captured until the completion of semantic triple transmission can be defined in a vector $\left[\boldsymbol{s}_1,\boldsymbol{a}_1, r_1, \boldsymbol{s}_2,\boldsymbol{a}_2, r_2,\ldots\right]$, which is referred as a trajectory. The cumulative reward of one complete semantic transmission trajectory is given as 
\begin{equation}\label{reward}
\begin{split}
R=&\left\{
\begin{aligned}
   \sum_{n=1}^N r_n+\gamma,\quad &\text{if}\ \mathbbm{1}_{\left\{E_A^N\leqslant\gamma\right\}},\\
-\eta,\quad &\text{otherwise},\\
\end{aligned}
\right.
\\=&\left\{
\begin{aligned}
   E_U^N-E_A^N+\gamma,\quad &\text{if}\ \mathbbm{1}_{\left\{E_A^N\leqslant\gamma\right\}},\\
-\eta,\quad &\text{otherwise},
\end{aligned}
\right.
\end{split}
\end{equation}
where $\eta>0$ is a constant penalty factor. Here, to encourage the agent to improve system privacy, we introduce $\gamma$ to give higher reward for private semantic transmission. In particular, $E_U^N$ in (\ref{reward}) can be taken as semantic transmission quality reward term and the term $\gamma-E_A^N$ can be used as an additional reward from the perspective of system privacy. From (\ref{reward}), we can see that when the transmission is not private, the cumulative reward will be negative, i.e., $-\eta$. Otherwise (i.e., $\mathbbm{1}_{\left\{E_A^N\leqslant\gamma\right\}}$), the cumulative reward will be non-negative, i.e. $\gamma-E_A^N\geqslant0$. In other words, this cumulative reward will drive the agent to avoid semantic triple leakage for additional reward.

 \item{\emph{Deterministic Policy}}: The deterministic policy is a mapping from a given state $\bm{s}_n$ to a deterministic action $\bm{a}_n$. The policy is implemented by the DNN based function approximator, i.e., actor network parameterized by $\bm{\phi}$, which establishes the relation between the mutual importance distribution of all semantic triples, the GNT of the user, and the semantic privacy. Compared to the stochastic policy that is the conditional probability of the agent choosing an action in a given state, the deterministic policy can converge with less samples by efficiently policy gradient estimation \cite{david2014dpg}. Specifically, the deterministic policy of the agent taking action $\bm{a}_n$ in a given state $\bm{s}_n$ can be expressed as $\bm{\pi}_{\bm{\phi}}\left(\bm{s}_n\right) = \bm{a}_n$.

\item{\emph{Q Value function}}: The Q value function $Q_{\bm{\theta}}\left(\bm{s}_n,\bm{a}_n\right)$ of the server is approximated by a DNN parameterized by $\bm{\theta}$, i.e. a critic network, which is used to estimate the expected future reward at state $\bm{s}_n$, given action $\bm{a}_n$. The input of this critic network is a state-action pair and the output is expected future reward. However, as studied in \cite{td3}, the output of such critic network can be overestimated. In particular, the local-optimal actions yielding high estimated Q value will be assigned with high probability during action exploration within traditional deep Q learning based RL algorithms. Such local-optimal actions will dominate the estimation of expected future reward, i.e., Q value, and will stay overestimated, which disturbs the action exploration and keeps the agent policy local-optimal. To solve such challenge, we merge the concept of clipped double Q learning, delayed policy update, and prioritized sampling into the training process of the PS-TD3 algorithm. 
\vspace{-0.1cm}
\end{itemize}

\subsection{Training of PS-TD3 Algorithm}
 Next, we will start the introduction of PS-TD3 training process with detailed explanation of the general training process of actor critic RL algorithm. The implementation of clipped double Q learning, the delayed policy update, and the prioritized sampling within this training process will be explained later.
 
\begin{figure}
\centering
\setlength{\belowcaptionskip}{-0.05cm}
\includegraphics[width=7cm]{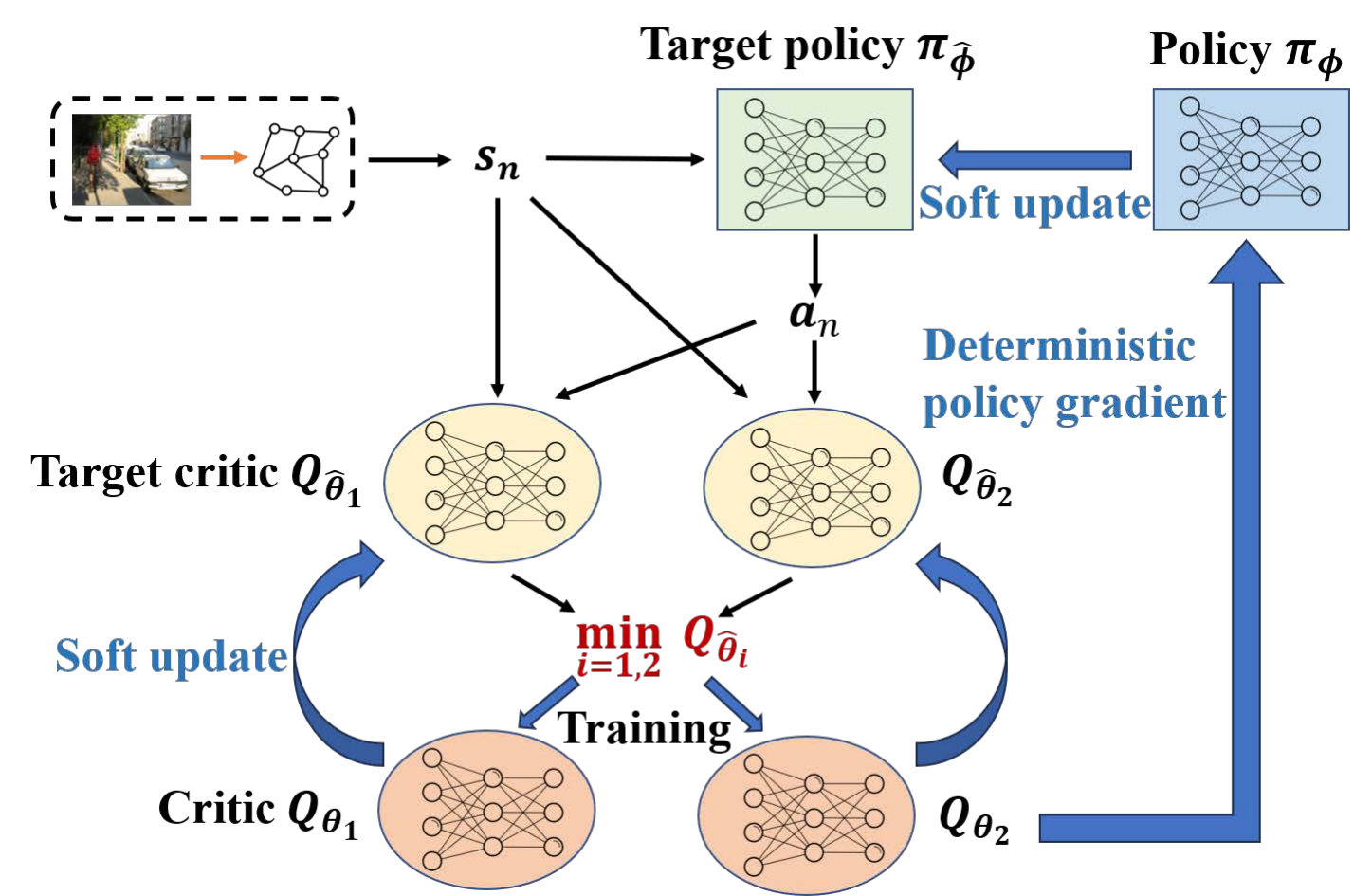}
\caption{The training process of the proposed PS-TD3 algorithm.}
\label{fig4}
\vspace{-0.2cm}
\end{figure} 
Generally, starting from randomly initialized policy and critic networks, the training of RL algorithms includes two stages: \subsubsection{Interaction stage for experience collection} The server feeds its current environmental observation (i.e. states) into the policy network, which outputs deterministic action choices on the transmission time slot selection and transmit power control for each triple. Then, the reward is calculated based on the received and the eavesdropped semantic information of the user and the attacker. After a number of interactions between the agent and environment, a set of transitions that consist of state, action, reward, and next state, i.e., $\left(\bm{s}_n,\bm{a}_n,r_n,\bm{s}_{n+1}\right)$, will be collected as historical experience and stored into a replay buffer ${\cal T}$.

\subsubsection{Experience replay stage for networks update} In this stage, the agent will randomly sample a series of transitions from the replay buffer to update actor and critic networks. Specifically, the critic network $Q_{\bm{\theta}}$ will be updated based on the temporal difference (TD) based loss function 
\begin{equation}\label{TD}
J(\bm{\theta}) = \frac{1}{N}\sum_{n=1}^N\left[y_n-Q_{\bm{\theta}}\left(\bm{s}_n,\bm{a}_n\right) \right]^2,
\end{equation}
where $n$ is the index of transition, $y_n=r_n+Q_{\hat{\bm{\theta}}}\left(\bm{s}_{n+1},\bm{a}_{n+1}\right)$ is the learning target with $\bm{s}_{n+1}$ being the next state and $\bm{a}_{n+1}$ being next action. Here, the estimated Q value of the next state-action pair is calculated by the target Q function $Q_{\hat{\bm{\theta}}}\left(\cdot\right)$ that is softly updated by $\hat{\bm{\theta}}\xleftarrow{} \tau\bm{\theta}+\left(1-\tau\right)\hat{\bm{\theta}}$ with $\tau$ being temperature parameter that adjusts updating rate. Then, based on updated critic network, the policy network can be updated by the deterministic policy gradient as in
\begin{equation}\label{dpg}
\begin{split}
    &\nabla_{\bm{\phi}} J\left(\bm{\phi}\right)=\\&\frac{1}{N}\sum_{n=1}^N \nabla_{\bm{a}_n}Q_{\bm{\theta}}\left(\bm{s}_n,\bm{a}_n\right)\mid_{\bm{a}_n=\pi_{\bm{\phi}}\left(\bm{s}_n\right)}\nabla_{\bm{\phi}}\pi_{\bm{\phi}}\left(\bm{s}_n\right),    
\end{split}
\end{equation}
where $\bm{\phi}$ is the vector of policy network parameters. 

This two stages will be implemented at the server alternately until a convergence is reached. The server will keep using the converged policy to make decisions on the transmission time slot and transmit power management for the covert semantic communication. As we mentioned in the former section, traditional actor-critic RL algorithms can suffer from overestimation of Q value. In what follows, we will treat this problem with the clipped double Q learning technique, as shown in Fig.~\ref{fig4}, and will also merge the concept of delayed policy update and prioritized sampling in this two stage training process for stabilized and high time efficient training performance. 

\begin{itemize}
\item {\textbf{Relieved overestimation with clipped double Q learning.}} With clipped double Q learning, we construct two critic networks $Q_{\bm{\theta}_1}$ and $Q_{\bm{\theta}_2}$, and only the smaller output Q value of these two target critic networks at each step will be chosen for TD calculation in (\ref{TD}). In other words, the learning target $y_n$ is reconstructed as 
\begin{equation}\label{yn}
    y_n=r_n + \min_{i\in\{1,2\}}Q_{\hat{\bm{\theta}}_i}\left(\bm{s}_{n+1},\bm{a}_{n+1}\right),
\end{equation}
where $\hat{\bm{\theta}}_i$ is the parameter of the target critic network $i\in\left\{1,2\right\}$. In this way, the experience replay stage is revised to be more conservative on taking dominating actions into the update of critic networks, so as to avoid overestimation. 

\item {\textbf{Smoothed training process with delayed policy update.}} Since the training of critic and policy networks are interdependent in the two training stages, we can expect that the training of policy can be unstable with unstable Q value output by target critic networks. Hence, to train the policy with a Q value estimated with lower variance, we propose to I) lower the update frequency of the policy network compared to critic networks. After a fixed number of updates to the critic networks, the policy network will be updated; II) add a clipped Gaussian noise $\mu$ to each output action, so as to enhance the exploration of the semantic communication environment with more diverse interaction and smoother estimation of Q value distribution; and III) introduce an additional target policy network $\pi_{\hat{\bm{\phi}}}$ whose parameters $\hat{\bm{\phi}}$ are soft updated by $\hat{\bm{\phi}}\xleftarrow{} \tau\bm{\phi}+\left(1-\tau\right)\hat{\bm{\phi}}$ for further relieved overestimation. 

\item {\textbf{Boosted training speed with prioritized sampling.}} Since both of the used pessimistic Q value estimation and the delayed policy update is conservative learning strategy that may need more iteration to converge compared to traditional RL, we need to further expedite the training process.  In particular, we proposed to integrate a prioritized sampling method into the interaction stage of the training process to make fully use of the collected transitions. For standard interaction, the agent randomly samples a set of historical experience, i.e., transitions, from the replay buffer for RL training, considering the importance of each transition for training RL to be the same. However, in practice the contribution of different transitions can be different. For example, the transition with higher TD error, i.e., $\delta_n=y_n-Q_{\bm{\theta}}\left(\bm{s}_n,\bm{a}_n\right)$, will lead to a larger updating step. This is because the higher TD error
indicates that the corresponding transition is more surprising and unexpected. Hence, the agent can learn more about Q value estimation from these transitions \cite{per}. Therefore, we introduce a TD error related priority to improve the probability of sampling the transitions with high TD error. In particular, the probability of sampling transition with index $n$ can be given by 
\begin{equation}\label{pr}
\textbf{Pr}\left(n\right)=\frac{b_n}{\sum_{i}b_i}=\frac{\left|\delta_n\right|^\alpha}{\sum_i \left|\delta_i\right|^\alpha}
\end{equation}
where $b_n=\left|\delta_n\right|^\alpha$ is the priority factor of the transition, $\alpha$ is a constant to adjust priority. The impacts from the prioritized sampling will be trivial when we set a small $\alpha$. Especially, $\alpha=0$ represents that no priority is considered in sampling. By integrating (\ref{pr}) into the interaction stage, the agent is more likely to sample the transition with higher TD error, which speedup the convergence of the PS-TD3 algorithm. The specific training procedure including the clipped double Q learning and the prioritized sampling is summarized in \textbf{Algorithm 1}.
\end{itemize}
\begin{algorithm}[t]
\caption{PS-TD3 algorithm for solving problem (\ref{eq:optimal_problem}).}
\begin{algorithmic}[1]
\footnotesize
\STATE \textbf{Initialize:} Actor and critic networks parameters $\bm{\phi},\bm{\theta}_1,\bm{\theta}_2$,
Target networks parameters $\hat{\bm{\phi}}=\bm{\phi}, \hat{\bm{\theta}}_1=\bm{\theta}_1, \hat{\bm{\theta}}_2=\bm{\theta}_2$,
Target network update rate $\tau$, 
Replay buffer ${\cal T}$,
Priority adjustment parameter $\alpha$.
\FOR {$u = 1 \to U$}
\FOR {each environment step}
\STATE Choose action $\bm{a}_n=\pi_{\hat{\bm{\phi}}}\left(\bm{s}_n\right)+\mu$ based on current target actor network.
\STATE Calculate the step reward $r_n\left(\bm{s}_n,\bm{a}_n\right)$.
\STATE Calculate TD error $\delta_n=r_n+\min_{i\in\{1,2\}}Q_{\hat{\bm{\theta}}_i}\left(\bm{s}_{n+1},\bm{a}_{n+1}\right)-Q_{\bm{\theta}}\left(\bm{s}_n,\bm{a}_n\right)$ and priority factor $b_n=\left|\delta_n\right|^\alpha$.
\STATE Collect transition into the buffer\\ 
${\cal T}={\cal T}\cup\left\{\left(\bm{s}_n,\bm{a}_n,r_n\left(\bm{s}_n,\bm{a}_n\right),\bm{s}_{n+1}, b_n\right)\right\}$.
\ENDFOR
\FOR {each network update step}
\STATE Sample a batch of $N$ transitions by the priority $b_n$.
\STATE Update critic networks $Q_{\bm{\theta}_1},Q_{\bm{\theta}_2}$ by $\nabla_{\bm{\theta}_i} {J}\left(\bm{\theta}_i\right)$.
\IF{a fixed number updating of critics have been done}
\STATE Update actor network $\bm{\phi}$ by $\nabla_{\bm{\phi}} {J}\left(\bm{\phi}\right)$.
\STATE Update target actor network by $\hat{\bm{\phi}}\xleftarrow{} \tau\bm{\phi}+\left(1-\tau\right)\hat{\bm{\phi}}$.
\STATE Update target critic networks by $\hat{\bm{\theta}}_i\xleftarrow{} \tau\bm{\theta}_i+\left(1-\tau\right)\hat{\bm{\theta}}_i$.
\STATE Initialize the updating number of critic networks to 0.
\ENDIF
\ENDFOR
\ENDFOR
\end{algorithmic}
\label{algorithm_1}
\end{algorithm}
 \begin{table}\footnotesize
\setlength{\belowcaptionskip}{0pt}
\setlength{\abovedisplayskip}{-15pt}
\setlength{\tabcolsep}{5mm}{
\newcommand{\tabincell}[2]{\begin{tabular}{@{}#1@{}}#1.0\end{tabular}}
\renewcommand\arraystretch{1.2}
\caption[table]{{System Parameters}}
\label{p}
\centering
\begin{tabular}{|c|c|c|c|c|c|}
\hline
\!\textbf{Parameter}\! \!\!& \!\!\!\!\textbf{Value} &\! \textbf{Parameter} \!& \!\!\!\!\textbf{Value}\!\!\! \\
\hline
$N$ & 12 &  $W$ & 2 KHz\\
\hline
$\beta$ & 2 &  $\alpha$ & 2\\
\hline
$T$ & 200 ms & $B$ & 1  \\
\hline
$\sigma_U^2$ & -30 dBm  & $\sigma_A^2$ & -30 dBm \\
\hline
$p_S^\text{max}$ & 1 W & $p_J^\text{max}$ & 1 W  \\
\hline
$G$ & 8 & $\gamma$ & 0.5  \\
\hline
$\eta$ & 1 &  &    \\
\hline
\end{tabular}}
\vspace{-0.2cm}
\end{table}
\begin{figure}[t]
\centering
\setlength{\belowcaptionskip}{-0.05cm}
\includegraphics[width=7cm]{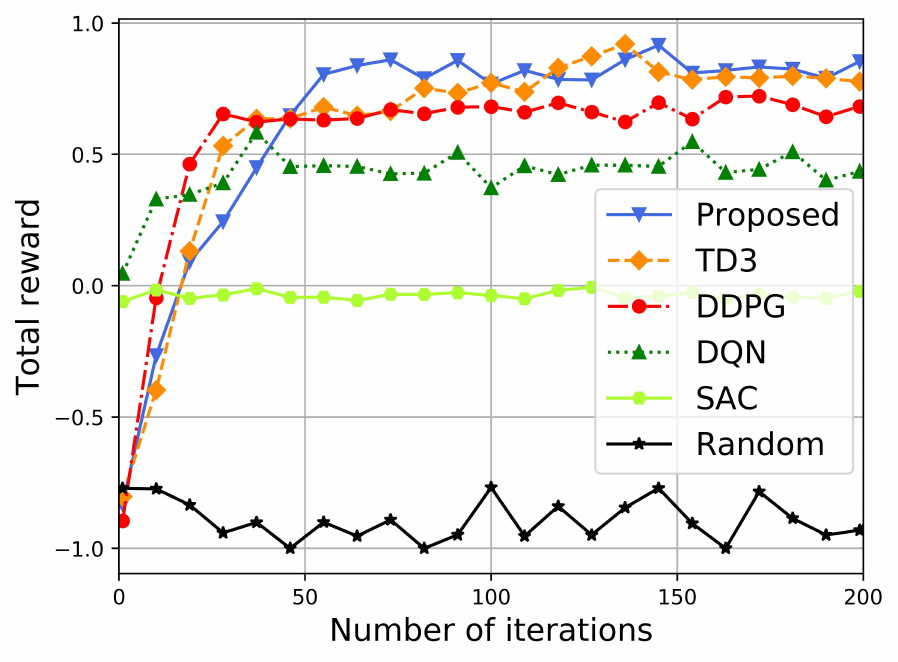}
\caption{Convergence of the proposed method.}
\vspace{-0.2cm}
\label{convergence1}
\end{figure}
\subsection{Complexity of the Proposed Algorithm}
 In this section, we analyze the complexity of the proposed PS-TD3 algorithm for covert semantic communication. The complexity of the PS-TD3 algorithm in training stage lies in actor and critic networks updating. In particular, the time-complexity of training each critic network that depends on the the number of hidden layers and neurons in each layer is given by
 \begin{equation}\label{complexity1}
{\cal O}\left(U_cN\sum_{j=1}^{{L_c}-1} \omega_j\omega_{j+1}+U_cN\left(K^2+2\right)\omega_1+U_cN\omega_{L_c}\right),
\end{equation}
where $U_c$ is the number of updating critic network, $N$ is the number of transitions in one batch, $w_j$ is the number of neurons in the $i$-th hidden layer, $L_c$ is the number of hidden layers in critic network, $K^2+2$ is the input dimension of critic network that equals to the sum of dimensions of state $\bm{s}_n$ and action $\bm{a}_n$. Since the critic network is used to evaluate a deterministic policy, the output is a Q value scalar, i.e., the output dimension is one. Hence, the computation complexity at last output layer is $\omega_{L_c}$. Similarly, we can obtain the time-complexity of training actor network, which is given by  
\begin{equation}\label{complexity2}
{\cal O}\left(U_vN\sum_{j=1}^{{L_v}-1} \omega_j\omega_{j+1}+U_vNK^2\omega_1+2U_vN\omega_{L_v}\right),
\end{equation}
where $U_v$ is the number of updating actor network, $L_v$ is the number of hidden layers in actor network. Compared to the critic network, the input dimension and output dimension of actor networks equal to the dimension of state and action, respectively. Furthermore, from the introduction of training the proposed algorithm, we can see that the training procedure of target networks is partial parameter coping. Hence, we can compute the time-complexity of updating target actor and critic networks are ${\cal O}\left(U_v\sum_{j=1}^{L_v}\omega_j+U_vK^2+2U_v\right)$ and  ${\cal O}\left(U_v\sum_{j=1}^{L_c}\omega_j+U_vK^2+2U_v\right)$, respectively. The proposed algorithm is trained offline. Therefore, after training, we only need to manage transmission time slot and power for private semantic communication by using actor network.

\begin{figure*}[t]
\centering  
\setlength{\belowcaptionskip}{-0.15cm}
\subfigure[Average GNT of the attacker.]{
\centering
\includegraphics[width=6cm]{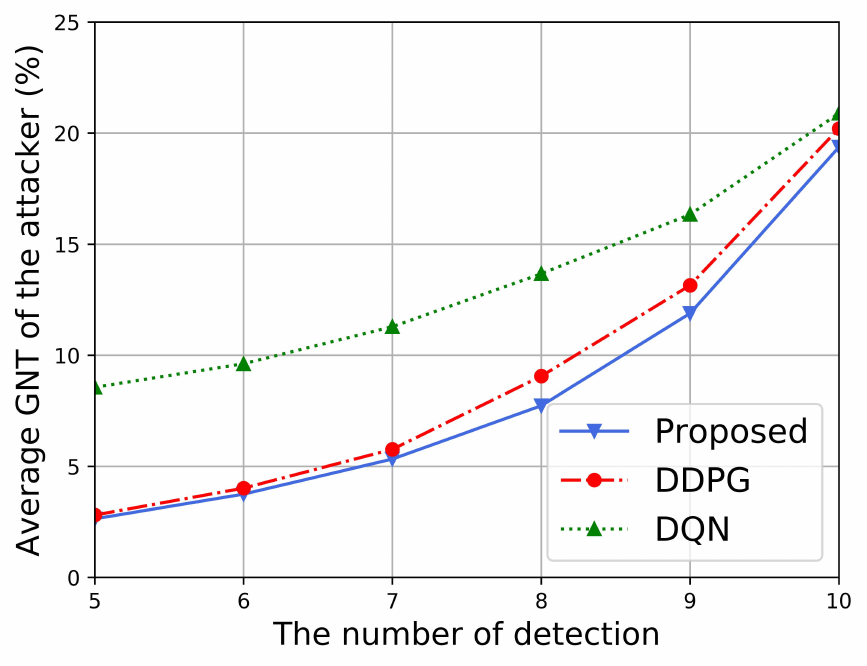}
}
\subfigure[Average GNT of the user.]{
\centering
\includegraphics[width=6cm]{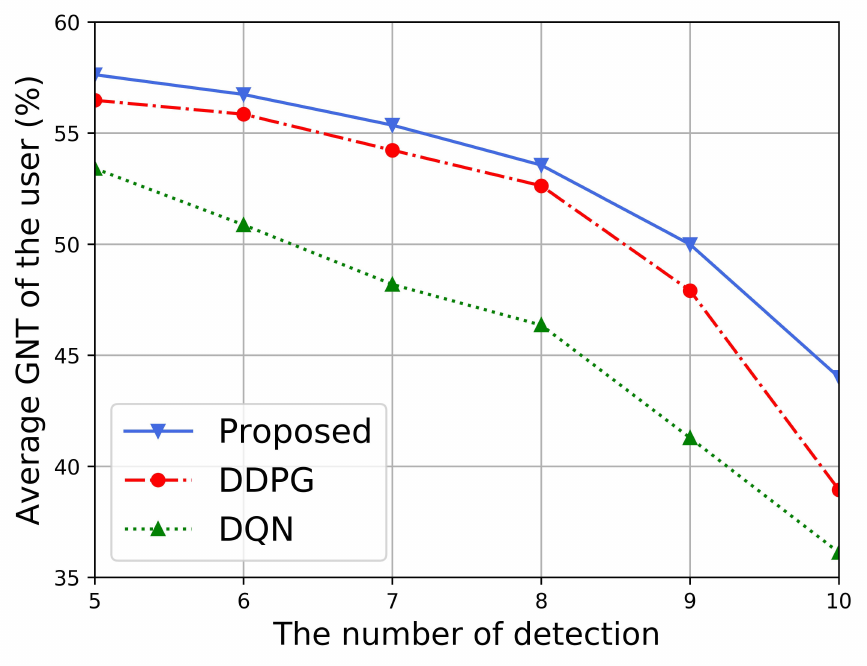}
}
\subfigure[Private transmission probability versus random method.]{
\centering
\includegraphics[width=6cm]{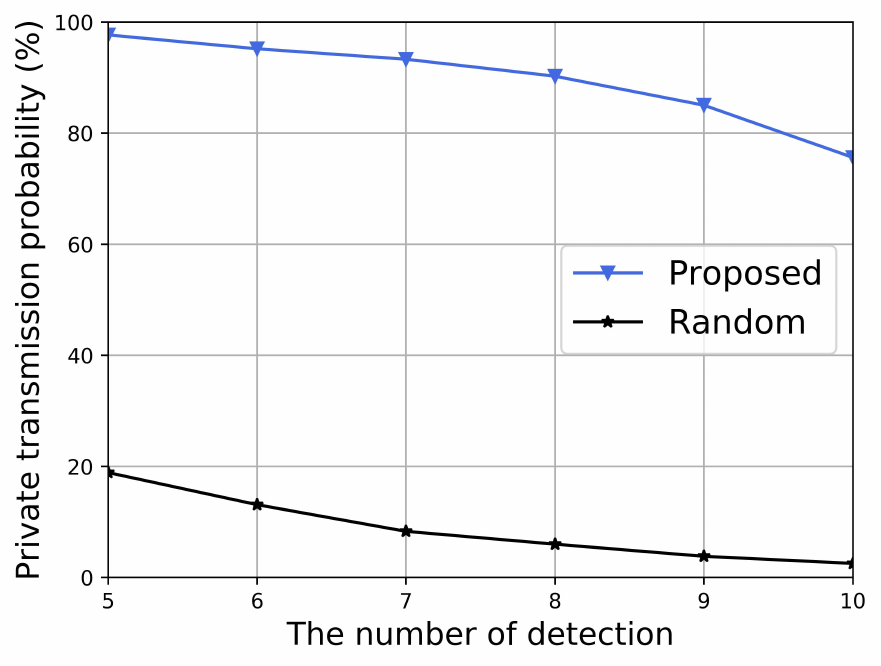}
}
\subfigure[Private transmission probability versus traditional RL methods.]{
\centering
\includegraphics[width=6cm]{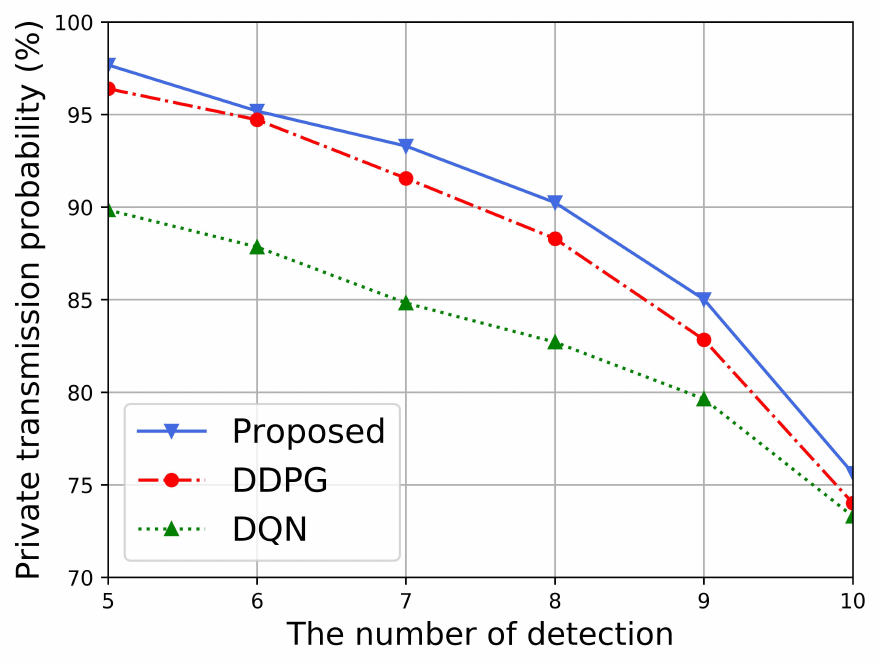}
}
\caption{Private semantic transmission performance as detection number varies.}
\label{detection-no}
\vspace{-0.2cm}
\end{figure*}

\begin{figure*}[t]
\centering  
\setlength{\belowcaptionskip}{-0.15cm}
\subfigure[Average GNT of the attacker.]{
\centering
\includegraphics[width=6cm]{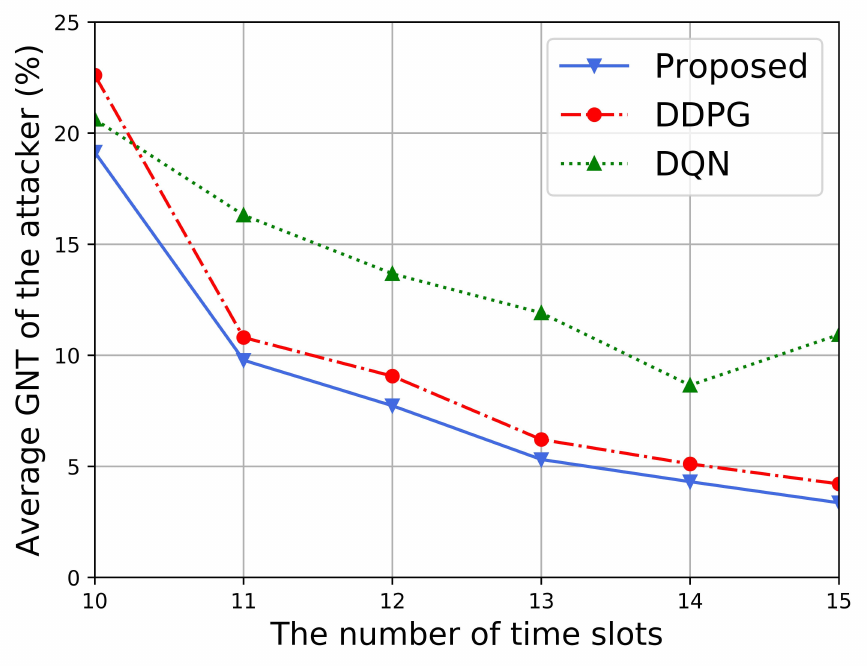}
}
\subfigure[Average GNT of the user.]{
\centering
\includegraphics[width=6cm]{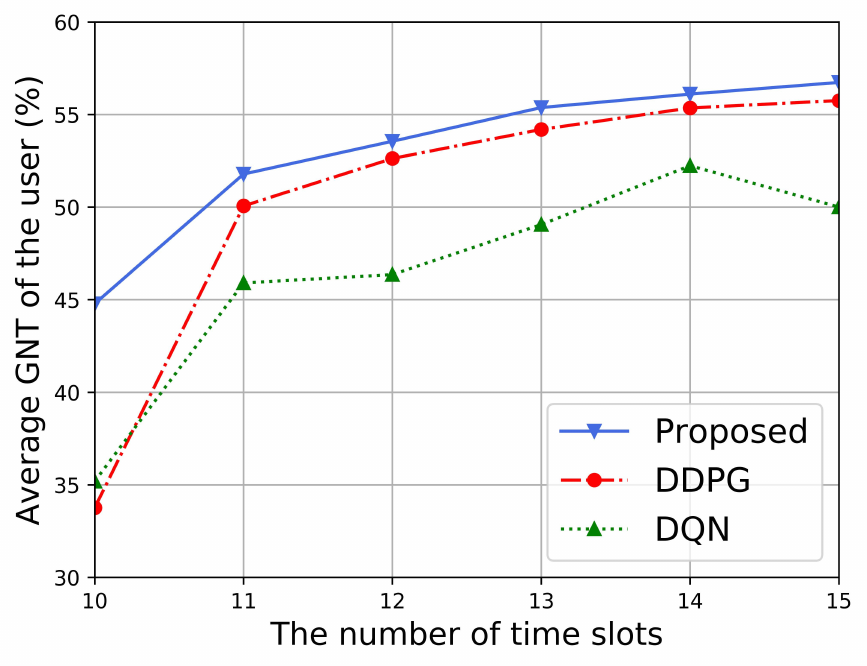}
}
\subfigure[Private transmission probability versus random method.]{
\centering
\includegraphics[width=6cm]{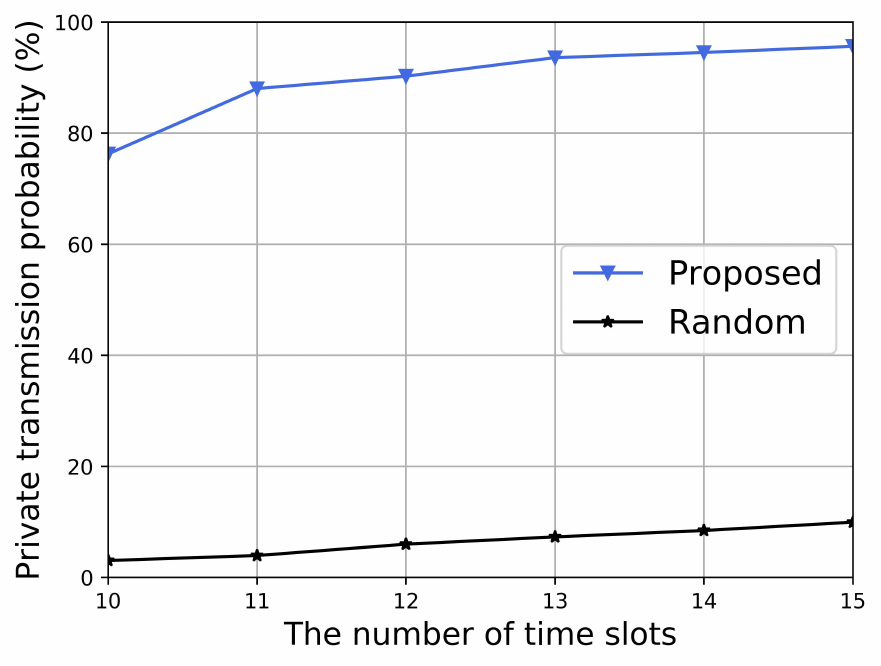}
}
\subfigure[Private transmission probability versus traditional RL methods.]{
\centering
\includegraphics[width=6cm]{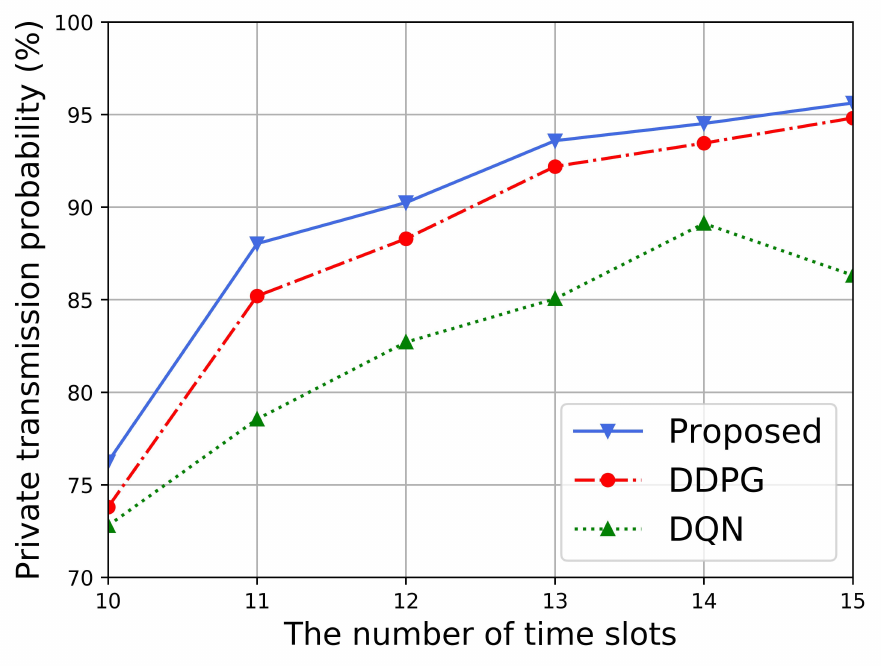}
}
\caption{Private semantic transmission performance as the number of transmission slots varies.}
\label{slot-no}
\vspace{-0.2cm}
\end{figure*}
\section{Simulation Results and Analysis}
In our simulations, we consider a circular wireless network where one server that locates at the center transmits image data to a randomly distributed user using semantic communication technique. At the same time, a randomly distributed attacker aims to eavesdrop the semantic information while a friendly jammer is deployed to protect the semantic transmission. The jammer constantly transmits jamming signal with random power. We assume the jamming power distribution remains the same during transmission. Hence, we trained and test the model under the same jamming power distribution. Here, we consider the attacker eavesdrops the transmission based on two kinds of power detectors in \cite{theoreticsecurity} and \cite{powerthreshold}. In particular, the detector in \cite{theoreticsecurity} is more accurate, which requires to know current ratio of transmit power and jamming signal power. The detector in \cite{powerthreshold} only requires to know the distribution of the jamming signal power. The detection number of the attacker is $G$. Other system parameters are listed in Table \ref{p}. We use the scene graph generation model in \cite{TDE} for semantic information extraction and the scene graph embedding module in \cite{justin2018sggen} for vectorization of semantic information. The training of RL algorithms require about 6000 semantic triple transmissions, equivalent to approximately 1000 images. For comparison purposes, we consider four baselines:

\begin{itemize}
\item{\emph{Deep Q learning network (DQN) algorithm}}. The DQN algorithm combines Q-learning with deep neural networks to approximate the optimal action-value function. The comparison between the proposed algorithm and DQN is to justify how the clipped Q function of the proposed method can improve the semantic communication privacy performance.

\item{\emph{Deep deterministic policy gradient (DDPG) algorithm}}. The DDPG algorithm combines actor critic architecture with deep neural networks to optimize both a deterministic policy and a Q value function. The comparison between the proposed algorithm and the DDPG algorithm is to justify how the double Q function and the delayed policy update of the proposed method can address the problem of  overestimating Q values. 

\item{\emph{Twin delayed deep deterministic policy gradient (TD3) algorithm}}. The TD3 algorithm is an advanced actor critic reinforcement learning method that introduces clipped double Q-learning to address overestimation bias and enhance stability. The comparison between the proposed algorithm and the TD3 algorithm is to justify how the introduced prioritized sampling of the proposed method can reduce the training iterations.

\item{\emph{Soft actor critic (SAC) algorithm}}. The SAC algorithm is the state-of-the-art actor critic reinforcement learning method that introduces entropy maximization learning to strengthen action exploration. The comparison between the proposed algorithm and the SAC algorithm is to justify how the deterministic policy of the proposed method can improve the training stability compared to stochastic policy.

\item{\emph{Random method that randomly chooses time slot and transmit power for semantic transmission}}. The comparison between the proposed algorithm and the random method is to justify how the proposed RL method enables the server to effectively optimize semantic transmission so as to improve system privacy level.
\end{itemize}

Figure \ref{convergence1} shows how the total reward of all considered solutions changes as the number of learning iterations varies. From Fig.~\ref{convergence1}, we see that the proposed deterministic policy of based PS-TD3 algorithm can achieve stable convergence compared to stochastic policy based SAC algorithm. In Fig.~\ref{convergence1}, we can also observe that compared with DQN and DDPG methods, the proposed algorithm can achieve about 77.8\% and 14.3\% improvement in terms of the reward, respectively, which can be directly reflected on higher system privacy and semantic quality of the received semantic information of the user. Such improvement stems from the fact that the use of clipped double Q function and delayed policy within the proposed algorithm encourages the agent to escape from local optimal actions and keep searching more advantageous ones. From the Fig.~\ref{convergence1}, we can also see that compared to the plain TD3 algorithm that requires about 80 iterations to converge, the proposed PS-TD3 algorithm reaches convergence after approximately 50 iterations. This is because the prioritized sampling mechanism introduced in the proposed method enables the agent to focus on more impactful transitions, thereby accelerating the training process. 

Figure \ref{detection-no} shows how the joint communication and privacy performance of the proposed algorithm varies with the attacker's detection number. Overall, from Fig.~\ref{detection-no}, we can see that as the detection number increases, the quality of semantic information eavesdropped by the attacker is improved such that the system's privacy crisis intensifies. This is because the server will tend to transmit more conservatively to avoid privacy leakage, e.g., lower transmit power, such that decreasing the transmission performance at the user. Meanwhile, Fig.~\ref{detection-no}a) shows that the proposed method keeps its average semantic similarity of eavesdropped semantic information lower than the other ones of baseline methods with the increasing detection numbers. Then, from Fig.~\ref{detection-no}b), we can see that the proposed method can acquire higher GNT of the user as the number of detection varies. This stems from the fact that the proposed algorithm enables the agent to aggressively search for the global optimal transmit power that can hide existence of transmission from frequent eavesdropping attacks. Finally, Fig.~\ref{detection-no}c) and Fig.~\ref{detection-no}d) shows how the probability of private semantic transmission achieved by all considered methods changes with the attacker's detection number. From Fig.~\ref{detection-no}c), we can see that the proposed method can significantly improve the privacy of semantic communication system compared to the random methods, which justify the effectiveness of the proposed RL based power control scheme. Meanwhile, from Fig.~\ref{detection-no}d), we observe that the proposed algorithm can achieve higher level of system privacy, compared to traditional RL methods. These gains stem from the fact that the combination of the clipped double Q function and delayed policy update in the proposed method enables the agent to aggressively search for the most effective transmission policy steadily without being mislead from Q value overestimation.
\begin{figure}[t]
\centering  
\setlength{\belowcaptionskip}{-0.15cm}
\subfigure[Average GNT of the user.]{
\centering
\includegraphics[width=6cm]{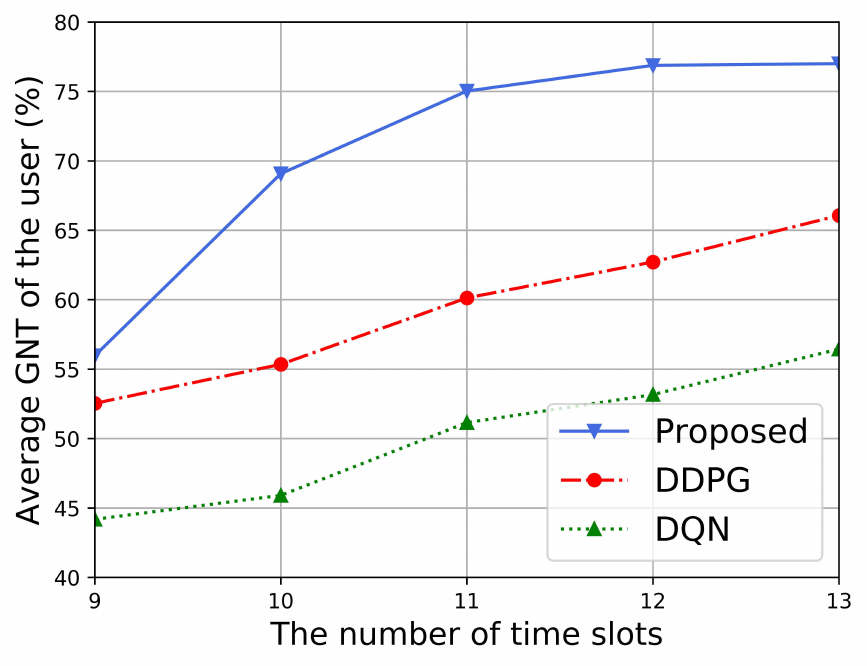}
}
\subfigure[Transmitted semantic triples percentage]{
\centering
\includegraphics[width=6cm]{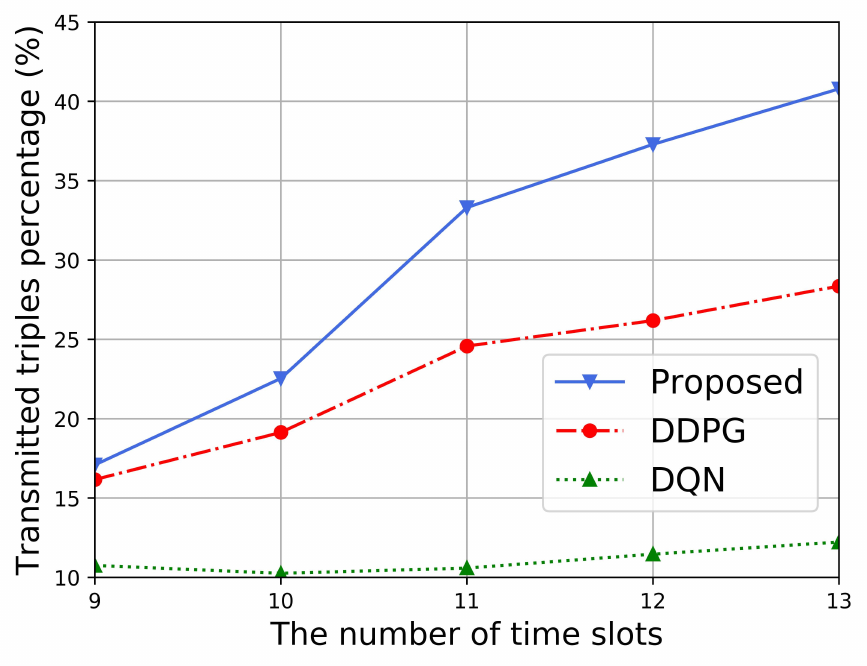}
}
\caption{Transmission performance confronts the power detector in \cite{powerthreshold}.}
\label{trans-rate}
\vspace{-0.2cm}
\end{figure}

Figure \ref{slot-no} shows how joint communication and privacy performance of the proposed PS-TD3 algorithm varies with the number of transmission time slots. In general, Fig.~\ref{slot-no} demonstrates that the semantic transmission become more private as the number of available transmission time slots increases, and the quality of the semantic information received by the user is also improved. This is because there will be more flexible choices for the server as the available time slots increasing. 
In Fig.~\ref{slot-no}a), we can observe that the proposed PS-TD3 algorithm can keep the quality of the eavesdropped semantic information in a low level, e.g., semantic similarity lower than 20\%. At the same time, from Fig.~\ref{slot-no}b), we can see that the proposed method can also acquire higher GNT of the received semantic information of the user as the number of transmission time slots changes. In Figs.~\ref{slot-no}c) and \ref{slot-no}d), we can also observe that the proposed method can significantly improve the system privacy compared to baseline methods. In particular, from Figs.~\ref{slot-no}a), \ref{slot-no}b), and \ref{slot-no}d), we can see that the DQN baseline method suffers non-trivial performance loss as the available transmission slots increases, i.e., enlarged action space. However, the proposed PS-TD3 algorithm can stably improve the semantic communication performance with the additional time slots. This is because the proposed PS-TD3 algorithm uses clipped double Q function to keep the server aggressively searching for global optimal solutions within the large action space, and uses the delayed updated deterministic policy to stabilize such searching. Finally, Figs.~\ref{slot-no}a), \ref{slot-no}b), and \ref{slot-no}d) also show a great performance loss at all considered RL baseline methods (i.e. DQN and DDPG) with reduced number of available time slots for transmission, while the proposed PS-TD3 algorithm can consistently achieve better performance in semantic quality and privacy as the number of time slots changes. The reason is that the accurately estimated Q value of the proposed RL method enable the agent to learn truly valuable actions. 
\begin{figure}[t]
\centering
\setlength{\belowcaptionskip}{-0.05cm}
\includegraphics[width=6cm]{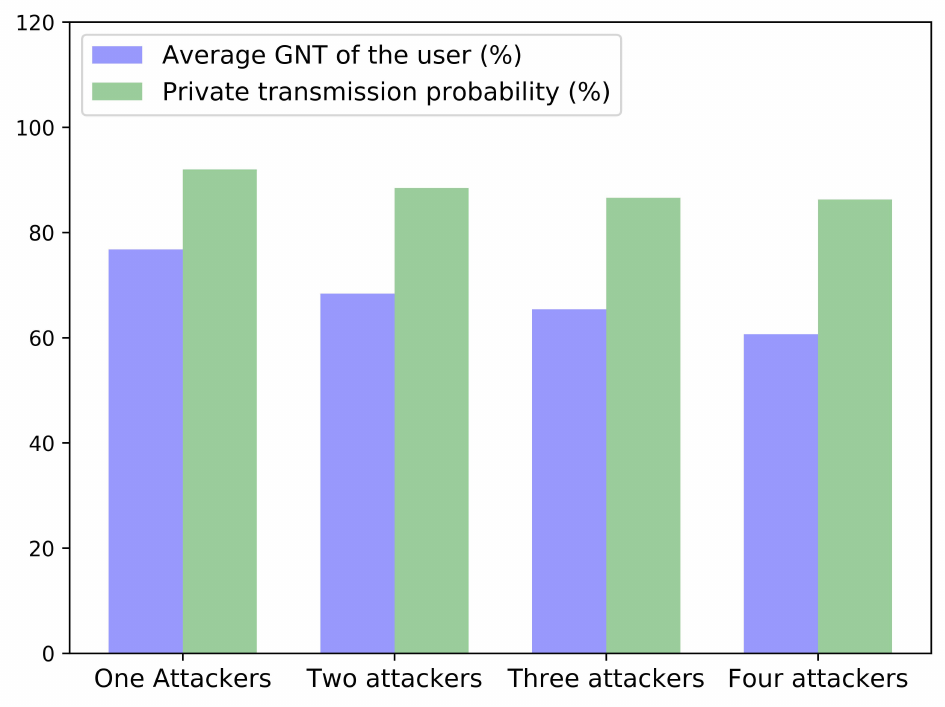}
\caption{Private semantic transmission performance in multi-attacker scenario}
\vspace{-0.2cm}
\label{multi-attacker}
\end{figure}

 Figure \ref{trans-rate} shows how the transmission performance varies as the number of available transmission slots increases, given that the attacker uses the power detector in \cite{powerthreshold}. Here, we note that, in this scenario, the privacy can be guaranteed by all the considered RL methods (private transmission probability over 90\%). From Fig.~\ref{trans-rate}a), we can see that the average GNT of the user increases with the available transmission time slots. Specifically, compared to the other two RL baseline methods, the quality of the received semantic information of the proposed PS-TD3 algorithm is significantly higher. Figure \ref{trans-rate}b) shows how the percentage of transmitted semantic triples changes with the number of transmission slots. A higher percentage means that more semantic triples can be transmitted, such that higher quality of semantic information can be achieved. From Fig.~\ref{trans-rate}b), we can see the transmitted semantic triples percentage of the proposed algorithm increases faster than DDPG method and DQN method. This is due to fact that the proposed PS-TD3 algorithm enables the server to effectively take advantage of additional transmission time slots to support more semantic triples transmission when confronts to a less powerful attacker, so as to improve semantic information quality of the user.
\begin{figure}[t]
\centering  
\setlength{\belowcaptionskip}{-0.15cm}
\subfigure[Under different attacker location.]{
\centering
\includegraphics[width=6cm]{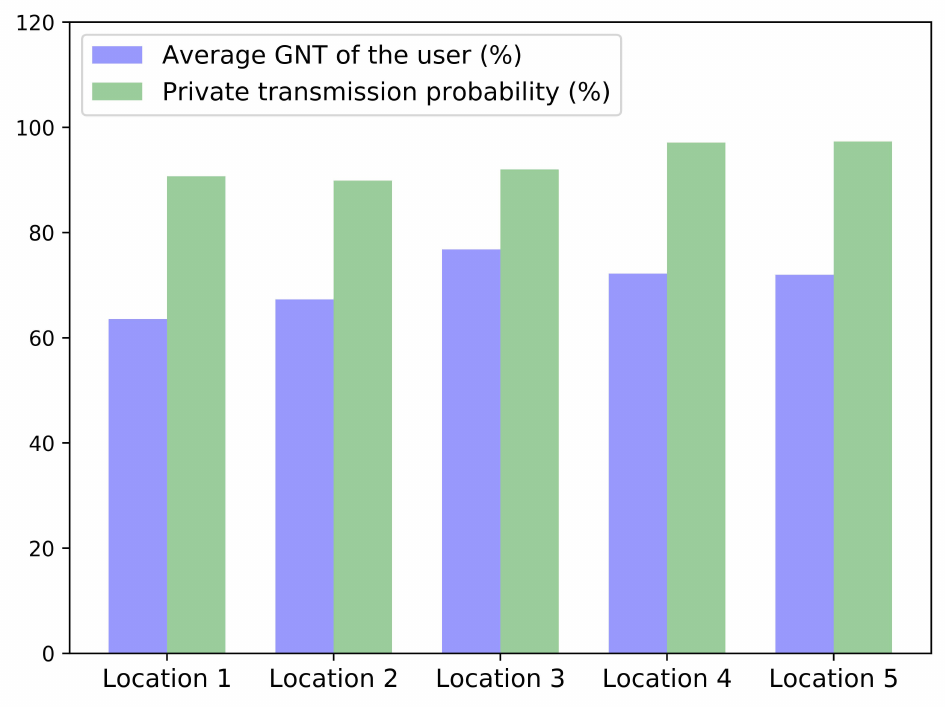}
}
\subfigure[Under different jamming power distribution.]{
\centering
\includegraphics[width=6cm]{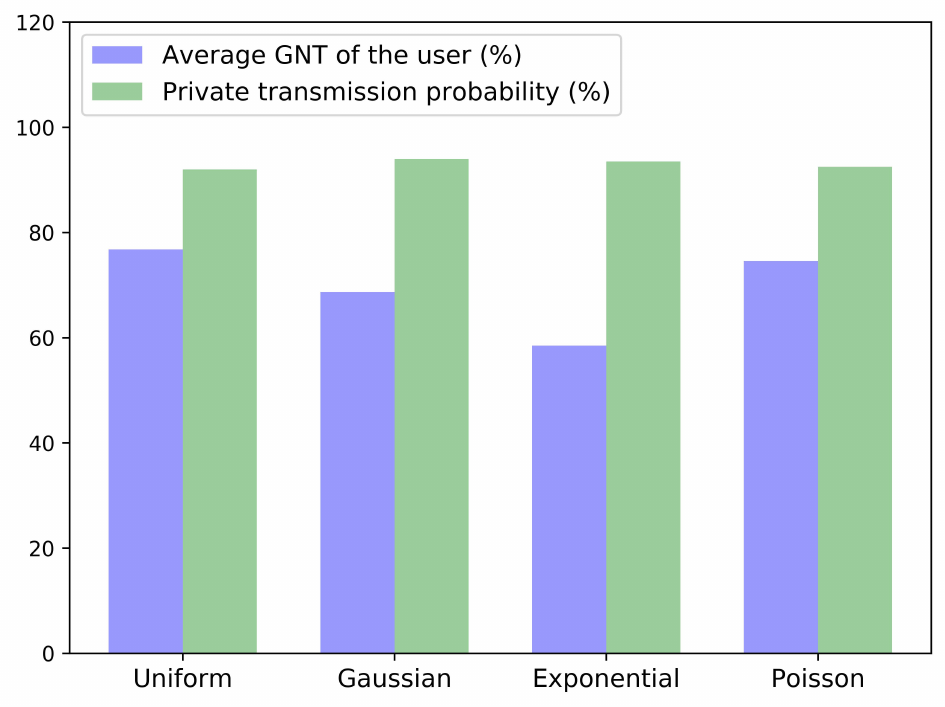}
}
\caption{Private semantic transmission performance under dynamic wireless environment.}
\vspace{-0.2cm}
\label{setting}
\end{figure}

Figure \ref{multi-attacker} shows the performance of the proposed algorithm in multi-attacker scenario. From Fig.~\ref{multi-attacker}, we see that the private transmission probability and the average GNT value of the user decrease as the number of the attacker increases. This is because the server have to choose transmit power more carefully to avoid eavesdropping by multiple attackers, which improve the difficulties in policy exploration of the proposed method. Meanwhile, in Fig.~\ref{multi-attacker}, we can also observe that the private transmission probability of the proposed algorithm remains 85\%, even in the presence of four attackers. This is because that the proposed method can still achieve relative private semantic transmission in multiple attackers scenario by stably exploring valuable actions.
\begin{figure}[t]
\centering  
\setlength{\belowcaptionskip}{-0.15cm}
\subfigure[Transmission slots arrangement of DQN method.]{
\centering
\includegraphics[width=7cm]{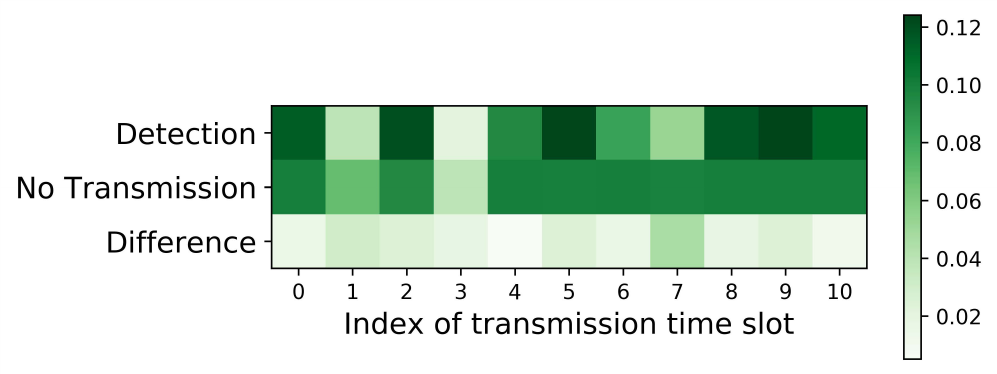}
}
\subfigure[Transmission slots arrangement of DDPG method.]{
\centering
\includegraphics[width=7cm]{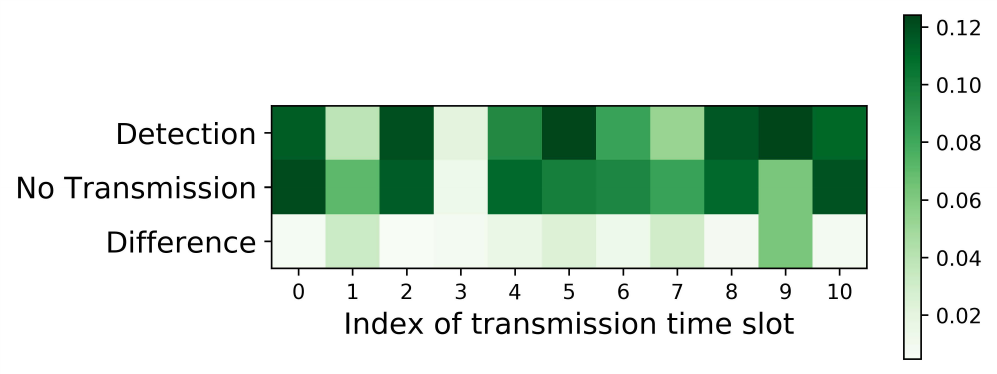}
}
\subfigure[Transmission slots arrangement of the proposed method.]{
\centering
\includegraphics[width=7cm]{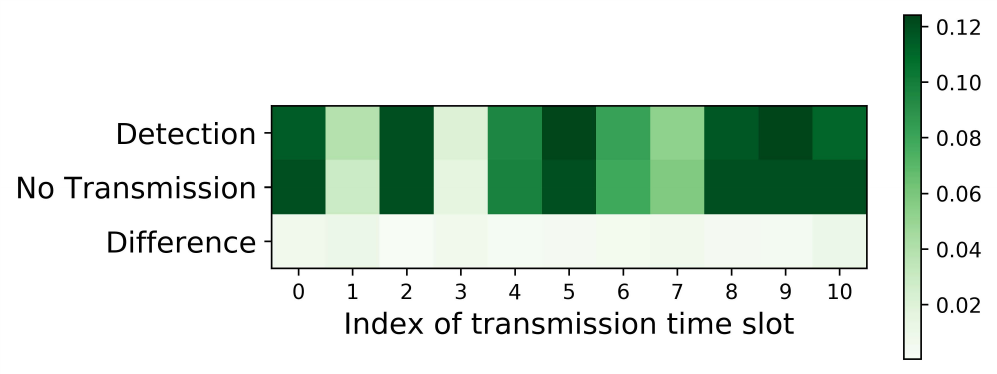}
}
\caption{Transmission time slot management.}
\label{slot-management}
\vspace{-0.2cm}
\end{figure}

Figure \ref{setting} shows the private semantic transmission performance of the proposed method under dynamic wireless environment including different attacker location and different jamming power distribution. In particular, Fig.~\ref{setting}a) shows how the attacker location affects the private transmission performance of the proposed method. Figure \ref{setting}b) shows how the jamming signal power distribution affects the private semantic transmission performance. From Figs.~\ref{setting}a) and b), we can see that no matter how we change the attacker's location and jamming signal power distribution, the proposed method can guarantee the system privacy and remain acceptable semantic transmission quality. 
This is because the maximized expected reward accounts the dynamics of the attacker locations and jamming power distributions.

Figure \ref{slot-management} shows the partial semantic triples transmission management at each time slot of the proposed method and baseline RL methods. The color of blocks represent the corresponding frequencies that computed based on 10,000 transmissions. In particular, as the frequency of detection at each time slot increases, the color of the first row in Figs.~\ref{slot-management}a), \ref{slot-management}b), and \ref{slot-management}c) changes from white to green. Similarly, the color of the second row in Fig.~\ref{slot-management} changes from white to green as the frequency of transmission at each time slot decreases (i.e., the frequency of no transmission increases). Then, the color changes of blocks in the third rows at Fig.~\ref{slot-management} represent the absolute value of difference between the blocks in the first row and the second row, i.e., the white block is corresponding to better privacy to avoid detection while the green block represents worse transmission protection. From Figs.~\ref{slot-management}a), b), and c), compared to baseline RL methods, we can see that the proposed PS-TD3 algorithm can enable the server to properly arrange partial semantic triples transmissions at each time slot to avoid detection.

\section{Conclusion}
In this paper, we have developed a novel covert semantic communication framework that consistently secures the semantic image transmission between a server and a user from eavesdropping attacks by jointly optimizing the semantic triple selection and transmit power control of each semantic transmission of the server. Within this framework, an independent friendly jammer is deployed to protect the semantic transmission without inter-device communication and cooperation for reduced spectrum abuse, energy costing, and information leakage. We have proposed a GNT metric to evaluate the communication performance and system privacy, and have cast this semantic triple selection and transmit power control problem in an optimization setting. We have also introduced a prioritized sampling assisted, twin delayed deep deterministic policy gradient RL solution to solve this non-convex problem in low computational and space complexity. Simulation results have shown that the proposed algorithm achieves high system privacy level and better quality of the received semantic information with fast convergence. 

\bibliographystyle{IEEEbib}
\def\baselinestretch{0.915}
\bibliography{ref}
\end{document}